\documentclass[11pt,fleqn,twoside,a4paper]{article}

\usepackage[german,english]{babel}
\usepackage{isreport-new}
\usepackage{amsmath}
\usepackage{latexsym}
\usepackage{babel}
\usepackage{amssymb}
\usepackage{times}
\usepackage{theorem}
\usepackage{xspace}
\usepackage{url}
\usepackage{alltt}
\usepackage{rotating}

\foreignreport

\title{\vspace{-2cm}Towards Automated Integration \\ of Guess and Check Programs 
\\ in Answer Set Programming:\\ A Meta-Interpreter and Applications}

\authornames{
Thomas Eiter \and Axel Polleres
}

\innertitle{~\\[-.5\baselineskip]
Towards Automated Integration of Guess and \\ Check Programs 
in Answer Set Programming:\\
A Meta-Interpreter and Applications
}

\author{
Thomas Eiter%
\affiliation{Institut f\"ur Informationssysteme, Technische Universit\"at Wien,
Favoritenstra\ss{}e\ 9-11, 1040 Wien, Austria.
\mbox{E-mail: eiter@kr.tuwien.ac.at.}}
\and
Axel Polleres%
\affiliation{\mbox{Institut f{\"u}r Informatik, Universit{\"a}t Innsbruck,
Technikerstra{\ss}e 21, A-6020 Innsbruck, Austria.}
\mbox{E-mail: axel.polleres@deri.org}}}

\acknowledgement{
This work was partially supported by FWF (Austrian Science Funds)
under projects \mbox{P14781} and \mbox{Z29-N04},  and 
Information Society Technologies
program of the European Commission, Future and Emerging Technologies
under the projects IST-2001-37004 (WASP) and IST-2001-33570 (INFOMIX).
}

\published{
\textbf{To appear in Theory and Practice of Logic Programming (TPLP)}. 
Part of the results in this paper are contained, in
  preliminary form, in the Proceedings of the 7th
International Conference on Logic Programming and Nonmonotonic
Reasoning (LPNMR~2004), Springer LNCS 2923, Ilkka Niemel{\"a} and Vladimir
Lifschitz (eds), pp.\ 100-113 and in the informal Proc.\ of the Joint Conference on Declarative Programming 
(APPIA-GULP-PRODE) 2003.}

\abstract{
Answer set programming (ASP) with disjunction offers a powerful tool for 
declaratively representing and
solving hard problems. Many $\NP$-complete problems can be encoded in
the answer set semantics of logic programs in a very concise and intuitive way, where the
encoding reflects the typical ``guess and check'' nature
of $\NP$ problems: The property is encoded in a way such that
polynomial size certificates for it correspond to
stable models of a program.  However, the problem-solving
capacity of full disjunctive logic programs (DLPs) is beyond
$\NP$, and captures a class of problems at the second level of the polynomial
hierarchy. While these problems also have a clear ``guess and check'' structure,
finding an encoding in a DLP reflecting this structure may sometimes be
a non-obvious task, in particular if the ``check'' itself is a
$\coNP$-complete problem; usually, such problems are solved by
interleaving separate guess and check programs, where the check is
expressed by inconsistency of the check program. In this paper, 
we present general transformations of head-cycle free (extended) disjunctive logic programs
into stratified and positive (extended) disjunctive logic programs based on
meta-interpretation techniques. The answer sets of the original and
the transformed program are in simple correspondence, and, moreover,
inconsistency of the original program is indicated by a designated
answer set of the transformed program.  Our transformations facilitate 
the integration of separate ``guess'' and ``check'' programs, which are often easy to
obtain, automatically into a single disjunctive logic program. Our
results complement recent results on meta-interpretation in ASP, and
extend methods and techniques for a declarative ``guess
and check'' problem solving paradigm through ASP.
\\*[\parskip]
}

\infsysrr{1843-04-01}

\date{January 2004}

\newcommand{\citeyear}[1]{\cite{#1}}

\newcounter{myenumctr}
\newenvironment{myenumerate}{\begin{list}{{\bf (\arabic{myenumctr})}\ }{\usecounter{myenumctr}
\setlength{\leftmargin}{0pt}
\setlength{\itemindent}{1.35\labelwidth}}}
{\end{list}}

\newcommand{\nop}[1]{}
\newcommand{\tuple}[1]{\ensuremath{\langle}#1\ensuremath{\rangle}}
\newcommand{\set}[1]{\ensuremath{\{#1\}}}

\newcommand{\dlv}{\texttt{\small DLV}\xspace}
\newcommand{\K}{\ensuremath{\mathcal{K}}\xspace}
\newcommand{\dlvk}{\ensuremath{\dlv^{\K}}\xspace}
\newcommand{\smodels}{{\sc Smodels}\xspace}
\newcommand{\gnt}{{\sc GnT}\xspace}

\def\Nat{\mathbb{N}}

\newcommand{\trP}{\ensuremath{tr}\xspace}
\newcommand{\tr}[1]{\ensuremath{\trP(#1)}\xspace}
\newcommand{\trOpt}[1]{\ensuremath{\trP_{Opt}(#1)}\xspace}

\newcommand{\HB}[1][ ]{\ensuremath{\mathit H\!b}\xspace} 

 \newcommand{\NP}{\mathrm{NP}}
 \newcommand{\coNP}{\textrm{co-}\NP}
 
 \newcommand{\SigmaP}[1]{{\Sigma}_{#1}^{P}}

\newcommand{\head}[1]{\ensuremath{H(#1)}}
\newcommand{\body}[1]{\ensuremath{B(#1)}}
\newcommand{\pbody}[1]{\ensuremath{B^+(#1)}}
\newcommand{\nbody}[1]{\ensuremath{B^-(#1)}}

\newcommand{\inlinek}[1]{\ensuremath{\mathtt{#1}}}
\newcommand{\tneg}{\ensuremath{\neg}}

\newcommand{\dneg}[1]{\ensuremath{\mathtt{not}\ }#1}
\newcommand{\naf}{\ensuremath{\mathtt{not}}\xspace}

\newcommand{\derives}{\mbox{\texttt{:\hspace{-0.15em}-}}\xspace}

\newcommand{\kuneq}{\ensuremath{\mathtt{\,!\!\!=}\,}}
\newcommand{\vel}{\mbox{\texttt{v}\xspace}}

\newenvironment{changemargin}{%
 \begin{list}{}{%
  \setlength{\topsep}{0pt}%
  \setlength{\leftmargin}{2em}
  \setlength{\listparindent}{0pt}%
  \setlength{\itemindent}{0pt}%
  \setlength{\parsep}{0pt}%
 }%
\item[]}{\end{list}}

\newtheorem{definition}{Definition}
\newtheorem{theorem}{Theorem}
\newtheorem{proposition}{Proposition}
\newtheorem{lemma}{Lemma}
\newtheorem{example}{Example}

\def\endproof{\ifhmode\nobreak\qed\par\fi\medskip} 
\newcommand{\qed}[0]{\hspace*{0mm}\hfill $\Box $\vspace{1.3mm}}

\begin{document}

\maketitle  
\sloppy
\newpage 

\pagenumbering{Roman}

\vspace*{\baselineskip}

\tableofcontents

\newpage 
\pagenumbering{arabic}

\section{Introduction}

Answer set programming (ASP)
\cite{asp-2001,gelf-lifs-91,lifs-2002,mare-trus-99,niem-99}, also
called A-Prolog \cite{bald-gelf-2003,bara-2002,gelf-2002}, is widely
proposed as a useful tool for solving problems in a declarative
manner, by encoding the solutions to a problem in the answer sets of a
normal logic program. By well-known complexity results, in this way
all problems with complexity in $\NP$ can be expressed and solved
\cite{schl-95,mare-remm-2003}; see also \cite{dant-etal-01}.

A frequently considered example of an $\NP$-complete problem which can
be elegantly solved in ASP is Graph-3-Colorability, i.e., deciding
whether some given graph $G$ is 3-colorable. It is an easy exercise in ASP
to write a program which determines whether a graph is 3-colorable. A
straightforward encoding, following the ``Guess and Check''
\cite{eite-etal-2000c,leon-etal-2002-dlv} respectively
``Generate/Define/Test'' approach \cite{lifs-2002}, consists of two
parts:
\begin{itemize}
    \item  A ``guessing'' part, which assigns nondeterministically each
node of the graph one of three colors:
{\small\begin{quote}
$\mathtt{col(red,X)\ \vel\ col(green,X)\ \vel\ col(blue,X)\ \derives\ node(X).}$
\end{quote}}  
\item and a ``checking'' part, which
tests whether no adjacent nodes have the same color:
{\small\begin{quote}
$\mathtt{\derives edge(X,Y),\ col(C,X),\ col(C,Y).}$  
\end{quote}}  
\end{itemize}
Here, the graph $G$ is represented by a set of facts \texttt{node($x$)} and
\texttt{edge($x,y$)}. Each legal 3-coloring of $G$ is a
polynomial-size ``proof'' of its 3-colorability, and such a given
proof can be validated in polynomial time. Furthermore, the answer sets
of this program yield all legal 3-colorings of the graph $G$.

However, we might encounter situations in which we want to express a
problem which is complementary to some $\NP$ problem, and thus belongs
to the class $\coNP$. It is widely believed that in general, not all
problems in $\coNP$ are in $\NP$, and hence that it is not always the
case that a polynomial-size ``proof'' of a $\coNP$ property $P$ exists
which can be verified in polynomial time. For such problems, we thus
can not write a (polynomial-size propositional) normal logic program
in ASP which guesses and verifies in its answer sets possible
``proofs'' of $P$.  One such property, for instance, is the
$\coNP$-complete property that a given graph is {\em not}
3-colorable. However, this and similar properties $P$ can be dually
expressed in ASP in terms of whether a normal logic program
(equivalently, a head-cycle free disjunctive logic program
\cite{bene-dech-94}) $\Pi_P$ has no answer set if and only if the
property $p$ holds.

Properties that are $\coNP$-complete often occur within the context of
problems that reside in the class $\SigmaP{2}$, which is above $\NP$
in the polynomial time hierarchy~\cite{papa-94}. In particular, the
solutions of a $\SigmaP{2}$-complete problem can be typically singled
out from given candidate solutions by testing a $\coNP$-complete
property.  Some well-known examples of such $\SigmaP{2}$-complete
problems are the following ones, which will be further detailed in
Section~\ref{sec:applications}:

\begin{description}

\item[Quantified Boolean Formulas:] Evaluating a Quantified Boolean
formula (QBF) of the form $\exists X\forall Y\Phi(X,Y)$, where
$\Phi(X,Y)$ is a disjunctive normal form over propositional variables
$X\cup Y$.  Here, a
solution is a truth value assignment $\sigma$ to the variables $X$
such that the formula $\forall Y\Phi(\sigma(X),Y)$ evaluates to true,
i.e., $\Phi(\sigma(X),Y)$ is a tautology.  Given a candidate solution
$\sigma$, the $\coNP$-complete property to check here is whether
$\Phi(\sigma(X),Y)$ is a tautology.

\item[Strategic Companies:] Computing strategic companies sets
\cite{eite-etal-2000c,leon-etal-2002-dlv}.  Roughly, here the problem
is to compute, given a set of companies $C$ in a holding, a minimal
subset $S \subseteq C$ which satisfies some constraints concerning the
production of goods and control of companies. Any such set is called
strategic; Given a candidate solution $S$ which satisfies the
constraints, the $\coNP$-complete property to check here is the
minimality, i.e., that no set $S' \subset S$ exists which also
satisfies the constraints.

\item[Conformant Planning:] Computing conformant plans under
incomplete information and nondeterministic action effects. Here the
problem is to generate from a description of the initial state $I$,
the planning goal $G$, and the actions $\alpha$ and their effects a sequence of actions (a
plan) $P = \alpha_1,\ldots,\alpha_n$ which carries the agent from the
initial state to a goal-fulfilling state under all contingencies, i.e.,
regardless of the precise initial state and how non-deterministic
actions work out. Given a candidate solution in terms of an {\em
optimistic}\/ plan $P$, which works under {\em
 some} execution \cite{eite-etal-2001e}, the property to check is
whether it works under {\em all }\/ executions, i.e., whether it is
conformant \cite{gold-bodd-96}. The latter problem is in $\coNP$,
provided that executability of actions is polynomially
decidable, cf.~\cite{eite-etal-2001e,turn-2002}. 
\end{description}

This list can be extended, and further examples can be found, e.g., in
\cite{eite-etal-97f,eite-etal-2002-tplp,grec-etal-2001,saka-inou-2003}.

The problems described above can be solved using ASP in a
two-step approach as follows:
\begin{enumerate}
\item Generate a candidate solution $S$ by means of a logic program
  $\Pi_{guess}$.
\item Check the solution $S$ by ``running'' another logic program
$\Pi_{check}$ (=$\Pi_p$)  on $S$, such that $\Pi_{check}\cup S$ has no
answer set if and only if $S$ is a valid  solution. 
\end{enumerate}

The respective programs $\Pi_{check}$ can be easily formulated (cf.\
Section~\ref{sec:applications}).

On the other hand, ASP with disjunction,
i.e.\ full extended disjunctive logic programming, allows one to formulate
problems in $\SigmaP{2}$ in a {\em single}\/ (disjunctive) program,
since this formalism captures the complexity class $\SigmaP{2}$, cf.\
\cite{dant-etal-01,eite-etal-97f}.  Hence, efficient ASP engines such
as \dlv~\cite{leon-etal-2002-dlv} or \gnt~\cite{janh-etal-2000} can be
used to solve such programs directly in a one-step approach. 

A difficulty here is that sometimes, an encoding of a problem in a
single logic program (e.g., for the conformant planning problem above)
may not be easy to find.  This raises the issue whether there exists
an (effective) possibility to {\em combine} separate $\Pi_{guess}$ and
$\Pi_{check}$ programs into a single program $\Pi_{solve}$, such that
this unified program computes the same set of solutions as the
two-step process outlined above. A potential benefit of such a
combination is that the space of candidate solutions might be reduced
in the evaluation due to its interaction with the checking
part. Furthermore, automated program optimization techniques may be
applied which consider both the guess and check part as well as the
interactions between them. This is not possible for separate programs.

The naive attempt of taking the union $\Pi_{guess} \cup \Pi_{check}$
unsurprisingly fails: indeed, each desired answer set of
$\Pi_{guess}$ would be eliminated by $\Pi_{check}$ (assuming that, in
a hierarchical fashion, $\Pi_{check}$ has no rules defining atoms
from $\Pi_{guess}$). Therefore, some program
transformation is necessary.  A natural question here is whether it is
possible to rewrite $\Pi_{check}$ to some other program
$\Pi'_{check}$ such that an integrated logic program $\Pi_{solve} = \Pi_{guess} \cup \Pi'_{check}$
is feasible, and, moreover, whether this can be done automatically. 

From theoretical complexity results about disjunctive logic programs
cf.\ \cite{dant-etal-01,eite-etal-97f}, one can infer that the program
$\Pi'_{check}$ should be truly disjunctive in general, i.e., not
rewritable to an equivalent non-disjunctive program in polynomial
time. This and further considerations (see Section~\ref{sec:trans})
provide some evidence that a suitable rewriting of $\Pi_{check}$ to
$\Pi'_{check}$ is not immediate.

In this paper, we therefore address this issue and present a generic
method for constructing the program $\Pi_{check}'$ by using a
meta-interpreter approach. In particular, we make the following
contributions:
\begin{myenumerate}
\item We provide a transformation $\trP(\Pi)$ from propositional
head-cycle-free \cite{bene-dech-94} (extended) disjunctive logic
programs (HDLPs) $\Pi$ to disjunctive logic programs (DLPs), which
enjoys the properties that the answer sets of $\tr{\Pi}$ encode the
answer sets of $\Pi$, if $\Pi$ has some answer set, and that
$\tr{\Pi}$ has a canonical answer set otherwise which is easy to
recognize.  The transformation $\tr{\Pi}$ is polynomial and modular in
the sense of \cite{janh-2000}, and employs meta-interpretation of
$\Pi$.

Furthermore, we describe variants and modifications of $\tr{\Pi}$
aiming at optimization of the transformation. In particular, we
present a transformation to positive DLPs, and show that in a precise
sense, modular transformations to such programs do not exist.


\item We show how to use $\trP(\cdot)$ for integrating separate guess
and check programs $\Pi_{guess}$ and $\Pi_{check}$, respectively, into a
single DLP $\Pi_{solve}$ such that the answer
sets of $\Pi_{solve}$ yield the solutions of the overall problem.

\item We demonstrate the method on the examples of QBFs, the Strategic
Companies problem, and conformant planning \cite{gold-bodd-96} under fixed
polynomial plan length (cf.\ \cite{eite-etal-2001e,turn-2002}). Our
method proves useful to loosen some restrictions of previous
encodings, and to obtain disjunctive encodings for more general
problem classes.

\item We compare our approach on integrating separate guess and check
programs experimentally against existing ad hoc encodings for QBFs and
Strategic Companies and also applying it to conformant planning, where
no such ad hoc encodings were known previously. For these experiments,
we use \dlv~\cite{leon-etal-2002-dlv}, a state-of-the-art Answer Set
engine for solving DLPs. The results which we obtained reveal
interesting aspects: While as intuitively expected, efficient ad hoc
encodings have better performance than the synthesized integrated
encodings in general, there are also cases where the performances
scale similarly (i.e., the synthesized encoding is within a constant
factor), or where even ad hoc encodings from the literature are
outperformed.
\end{myenumerate}

Our results contribute to further the ``Guess and Check'' resp.\
``Generate/Define/ Test'' paradigms for ASP, and fill a gap by
providing an automated construction for integrating guess and check
programs. They relieve the user from the burden to use sophisticated
techniques such as saturation, as employed e.g.\ in
\cite{eite-etal-97f,eite-etal-2000c,leon-etal-2001}, in order to
overcome the technical intricacies in combining natural guess and
check parts into a single program. Furthermore, our results complement
recent results about meta-interpretation techniques in ASP, cf.\
\cite{mare-remm-2003,delg-etal-01,eite-etal-2002a}.

The rest of this paper is organized a follows. In the next section, we
very briefly recall the necessary concepts and fix notation. After
that, we present in Section~\ref{sec:trans} our transformation
$\tr{\Pi}$ of a ``checking'' program $\Pi$ into a disjunctive logic
program. We start there with making the informal desirable properties
described above more precise, present the constituents of $\tr{\Pi}$,
the factual program representation $F(\Pi)$ and a meta-interpreter
$\Pi_{meta}$, and prove that our transformation satisfies the
desirable properties. Section~\ref{sec:optimizations} thereafter is
devoted to modifications towards optimization.  In
Section~\ref{sec:integration}, we show how to synthesize separate
guess and check programs to integrated encodings. Several applications
are considered in Section~\ref{sec:applications}, and experimental
results for these are reported in Section~\ref{sec:experiments}. The
final Section~\ref{sec:conclusion} gives a summary and presents issues
for further research.

\section{Preliminaries}
\label{sec:prelim}

We assume that the reader is familiar with logic programming and
answer set semantics, see \cite{gelf-lifs-91,asp-2001}, and only
briefly recall the necessary concepts.


A \emph{literal} is an atom $a(t_1, \ldots,t_n)$, or its negation
$\tneg{a(t_1, \ldots,t_n)}$, where ``\tneg{}'' is the strong negation
symbol, for which we also use the customary ``--'', in a function-free
first-order language (including at least one constant), which is
customarily given by the programs considered.  We write $|a| = |\tneg{a}| = a$ to
denote the atom of a literal.

Extended disjunctive logic programs (EDLPs; or simply programs) are
disjunctive logic programs with default (weak) and strong negation,
i.e., finite sets $\Pi$ of rules $r$
\begin{equation}
\label{stmt:lprule}
h_1 \vel\ \ldots\ \vel\ h_l\ \derives\ b_1,\ \ldots,\ b_m,\
\dneg{b_{m+1}},\ \ldots\ \dneg{b_n}.
\end{equation}
\noindent $l,m,n \geq 0$, where each $h_i$ and $b_j$ is a literal
and \dneg is weak negation (negation as failure). By $\head{r} =
\{h_1, \ldots, h_l\}$, $\pbody{r} = \{b_1,\ldots, b_m\}$,
$\nbody{r} = \{b_{m+1},\ldots, b_n\}$, and $\body{r} = \pbody{r}
\cup \nbody{r}$ we denote the head and (positive, resp.\
negative) body of rule $r$.
Rules with $|\head{r}|{=}1$ and $\body{r}{=}\emptyset$ are
called \emph{facts} and rules with $\head{r}{=}\emptyset$ are
called \emph{constraints}. For convenience, we omit ``extended'' in
what follows and refer to EDLPs as DLPs etc.

Literals (resp.\ rules, programs) are \emph{ground} if they are
variable-free. Non-ground rules (resp.\ programs)
amount to their \emph{ground instantiation},
i.e.,\ all rules obtained by substituting variables with constants from
the (implicit) language.

Rules (resp.\ programs) are \emph{positive}, if ``$\naf$'' does not occur
in them, and \emph{normal}, if $|\head{r}| \leq
1$. A ground program $\Pi$ is \emph{head-cycle free}
\cite{bene-dech-94}, if no literals $l\neq l'$ occurring in the same
rule head mutually depend on each other by positive recursion; $\Pi$
is stratified \cite{przy-89b,przy-91}, if no literal $l$ depends by recursion
through negation on itself (counting disjunction as positive recursion).


The {\em answer set semantics}~\cite{gelf-lifs-91} for DLPs is as
follows. Denote by $Lit(\Pi)$ the set of all ground literals for a
program $\Pi$. Consider first positive (ground) programs $\Pi$. Let $S
\subseteq Lit(\Pi)$ be a set of consistent literals. Such a set $S$
satisfies a positive rule $r$, if $\head{r} \cap S \not= \emptyset$
whenever $\pbody{r} \subseteq S$. An \emph{answer set} for $\Pi$ then
is a minimal (under $\subseteq$) set $S$ satisfying all rules.%
\footnote{We disregard a possible inconsistent answer set, which
is not of much interest for our concerns.}  To extend this definition to
programs with weak negation, the \emph{reduct} $\Pi^S$ of a program
$\Pi$ with respect to a set of literals $S$ is the set of rules
$$h_1\ \vel\ \ldots\ \vel\ h_l\ \derives\ b_1,\ \ldots,\ b_m$$
for all rules (\ref{stmt:lprule}) in $\Pi$ such that $S \cap \nbody{r} = \emptyset$.
Then $S$ is an {\em answer set} of $\Pi$, if $S$ is an answer set for $\Pi^S$.

There is a rich literature on characterizations of answer sets of DLPs
and restricted fragments; for our concerns, we recall here the
following characterization of (consistent) answer sets for HDLPs,
given by Ben-Eliyahu and Dechter~\citeyear{bene-dech-94}:

\begin{theorem}
\label{theo:hedlp}
 Given a ground HDLP $\Pi$, a consistent $S \subseteq Lit(\Pi)$ is an answer set
iff
\begin{enumerate}
\item $S$ satisfies each rule in $\Pi$, and
\item there is a function $\phi: Lit(\Pi) \mapsto \Nat$ such that for each
literal $l$ in $S$ there is a rule $r$ in $\Pi$ with
   \begin{enumerate}
   \item $\pbody{r} \subseteq S$
   \item $\nbody{r} \cap S = \emptyset$
   \item $l \in \head{r}$
   \item $S \cap (\head{r}\setminus \{l\}) = \emptyset$
   \item $\phi(l') < \phi(l)$ for each $l' \in \pbody{r}$
   \end{enumerate}
\end{enumerate}
\end{theorem}

We will use Theorem~\ref{theo:hedlp} as a basis for the transformation
$\tr{\Pi}$ in the next section.

\section{Meta-Interpreter Transformation}
\label{sec:trans}

As discussed in the Introduction, rewriting a given check program
$\Pi_{check}$ to a program $\Pi_{check}'$ for integration with a
separate guess program $\Pi_{guess}$ into a single program
$\Pi_{solve} = \Pi_{guess}\cup \Pi'_{check}$ can be difficult in
general. The problem is that the working of the answer set semantics,
to be emulated in $\Pi_{check}'$, is not easy to express there.

One difficulty is that for a given answer set $S$ of $\Pi_{guess}$, we
have to test the {\em non-existence} of an answer set of $\Pi_{check}$
with respect to $S$, while $\Pi_{solve}$ {\em should have an answer
set} extending $S$ to $\Pi'_{check}$ if the check succeeds. A
possibility to work around this problem is to design $\Pi'_{check}$ in
a way such that it has a dummy answer set with respect to $S$ if the
check of $\Pi_{check}$ on $S$ succeeds, and no answer set if the check
fails, i.e., if $\Pi_{check}$ has some answer set on $S$.  While this
may not look to be very difficult, the following observations suggest
that this is not straightforward.

Since $\Pi_{solve}$ may need to solve a $\SigmaP{2}$-complete problem,
any suitable program $\Pi'_{check}$ must be truly disjunctive in general, i.e.,
contain disjunctions which are not head-cycle free (assuming that no
head literal in $\Pi'_{check}$ occurs in $\Pi_{guess}$).  Indeed, if
both $\Pi_{guess}$ and $\Pi'_{check}$ are head-cycle free, then also
$\Pi_{solve} = \Pi_{guess} \cup \Pi'_{check}$ is head-cycle free, and
thus can only express a problem in $\NP$.

Furthermore, we can make in $\Pi_{check}'$ only limited use of default
negation on atoms which do not occur in $\Pi_{guess}$. The reason is
that upon a ``guess'' $S$ for an answer set of $\Pi_{solve} =
\Pi_{guess} \cup \Pi_{check}'$, the reduct $\Pi_{solve}^S$ is
$\naf$-free. Contrary to the case of $\Pi_{check}$ in the two-step
approach, it is not possibile to explicitly consider for a guess
$S_{guess}$ of an answer set of $\Pi_{guess}$ varying extensions $S =
S_{guess}\cup S'_{check}$ to the whole program $\Pi_{solve}$ which
activate different rules in $\Pi'_{check}$ (e.g.,\ unstratified
clauses $a \derives \naf\,b$ and $b\derives \naf\, a$ encoding a choice among
$a$ and $b$). Therefore, default negation in rules of $\Pi_{check}$
must be handled with care and might cause major rewriting as well.

These observations provide some evidence that a rule-rewriting
approach for obtaining $\Pi'_{check}$ from $\Pi_{check}$ may be
complicated. For this reason, we adopt at
a generic level a Meta-interpreter approach, in which the
$\coNP$-check modeled by $\Pi_{check}$ is ``emulated'' by a
minimality check for a positive DLP $\Pi_{check}'$.


\subsection{Basic approach}

The considerations above lead us to an approach in which the program
$\Pi_{check}'$ is constructed by the use of meta-interpretation
techniques \cite{mare-remm-2003,delg-etal-01,eite-etal-2002a}.  The
idea behind meta-interpretation is here that a program $\Pi$ is represented
by a set of facts, $F(\Pi)$, which is input to a fixed program
$\Pi_{meta}$, the meta-interpreter, such that the answer sets of
$\Pi_{meta} \cup F(\Pi)$ correspond to the answer sets of $\Pi$. Note
that the meta-interpreters available are normal logic programs
(including arbitrary negation), and can not be used for our purposes
for the reasons explained above.  We thus have to construct a novel
meta-interpreter which is essentially $\naf$-free, i.e.\ uses negation
as failure only in a restricted way, and contains disjunction.

Basically, we present a general approach to translate normal LPs and HDLPs 
into stratified disjunctive logic programs.
To this end, we exploit Theorem~\ref{theo:hedlp} as a basis for a
transformation $\tr{\Pi}$ from a given HDLP $\Pi$ to a DLP $\tr{\Pi} =
F(\Pi) \cup \Pi_{meta}$ such that $\tr{\Pi}$ fulfills the properties
mentioned in the introduction. More precisely, it will satisfy the
following properties:

\begin{description}
 \item[T0] $\tr{\Pi}$ is computable in time polynomial in the size of
 $\Pi$.
 \item[T1]
      Each answer set $S'$ of the transformed program $\tr{\Pi}$
      corresponds to an answer set $S$ of $\Pi$, such
      that $S = \{ l \mid \texttt{inS}(l) \in S'\}$ for some predicate
      \texttt{inS($\cdot$)}, provided $\Pi$ is consistent, and
      conversely, each answer set $S$ of $\Pi$ corresponds to some answer
      set $S'$ of  $\tr{\Pi}$ such that $S = \{ l \mid \texttt{inS}(l) \in S'\}$.
\item[T2] If the program $\Pi$ has no answer set, then $\tr{\Pi}$
      has exactly one designated answer set $\Omega$, which is easily
      recognizable.
\item[T3] The transformation is of the form $\trP(\Pi) = F(\Pi) \cup
  \Pi_{meta}$, where $F(\Pi)$ is a factual representation of $\Pi$ and
  $\Pi_{meta}$ is a fixed meta-interpreter.
\item[T4] $\tr{\Pi}$ is {\em modular} (at the syntactic level), i.e.,
$\tr{\Pi} = \bigcup_{r\in \Pi} \tr{r}$ holds. Moreover, $\trP(\Pi)$
returns a stratified DLP \cite{przy-89b,przy-91} which uses negation
only in its ``deterministic'' part.
\end{description}

Note that properties {\bf T0} -- {\bf T4} for $\tr{\cdot}$ are similar
yet different from the notion of polynomial faithful modular (PFM)
transformation by Janhunen~\citeyear{janh-2000,janh-2001}, which is a
function $Tr$ mapping a class of logic programs $\cal C$ to another
class $\cal C'$ of logic programs (where $\cal C'$ is assumed to be a
subclass or superclass of $\cal C$), such that the following three
conditions hold: (1) For each program $\Pi\in {\cal C}$, $Tr(\Pi)$ is
computable in polynomial time in the size of $\Pi$ (called {\em
polynomiality}), (2) the Herbrand base of $\Pi$, $\HB(\Pi)$, is
included in the Herbrand base of $Tr(\Pi)$, $\HB(Tr(\Pi))$ and the
models/interpretations of $\Pi$ and $Tr(\Pi)$, are in one-to-one
correspondence and coincide up to $\HB(\Pi)$ ({\em faithfulness}), and
(3) $Tr(\Pi_1\cup \Pi_2)$ = $Tr(\Pi_1)\cup Tr(\Pi_2)$ for all programs
$\Pi_1,\Pi_1$ in ${\cal C}$ and ${\cal C}'\subseteq {\cal C}$ implies
$Tr(\Pi)=\Pi$ for all $\Pi$ in ${\cal C}'$ ({\em modularity}).

Compared to PFM, also our transformation $\tr{\cdot}$ is polynomially
computable by {\bf T0} and hence satisfies condition 1). Moreover, by {\bf T4} and
the fact that stratified disjunctive programs are not necessarily
head-cycle free, it also satisfies condition 3). However, condition 2)
fails. Its first part, that $\HB(\Pi) \subseteq \HB(\tr{\Pi})$ and that
answer sets coincide on $Lit(\Pi)$ could be fulfilled by adding rules
$l\ \derives\ \texttt{inS}(l)$ for every $l \in Lit(\Pi)$); these
polynomially many rules could be added during input generation.
The second part of condition 2) is clearly in contradiction with {\bf
T2}, since for $\Omega$ never a corresponding answer set of $\Pi$ exists.
Moreover, condition {\bf T1} is a weaker condition than the
one-to-one correspondence between the answer sets of
$\Pi$ and $\tr{\Pi}$ required for faithfulness:
In fact, in case $\Pi$ has positive cycles, there might be several
possible guesses for $\phi$  for an answer
set $S$ of $\Pi$ in Theorem~\ref{theo:hedlp} reflected by
different answer sets of $\tr{\Pi}$.
We illustrate this by a short example:

\begin{example}\rm Let $\Pi$ be the program consisting of the following
four rules:
{\tt
\begin{tabbing}
\ \ \ \mbox{$r1:$}\ a\ \derives\ b. \quad  \mbox{$r2:$}\ b\ \derives\ a.
\quad  \mbox{$r3:$}\ a. \quad \mbox{$r4:$}\ b.
\end{tabbing}
}
\noindent Then, $\Pi$ has a single answer set $S = \set{\inlinek{a,b}}$, while
$\tr{\Pi}$ has two answer sets such that
$S_1 = \set{\inlinek{inS(a),inS(b), phi(a,b), \ldots}}$ and
$S_2 = \set{\inlinek{inS(a),inS(b), phi(b,a), \ldots}}$, intuitively
reflecting that here the order of applications of rules $r1$ and $r2$
does not matter, although they are cyclic.
\end{example}

We remind that the different properties of our transformation
$\tr{\cdot}$ and PFM transformations is not an accident but a
necessary feature, since we want to express nonexistence of certain
answer sets via the transformation, and not merely preserve the exact
semantics as targeted by PFM. Apart from this different objective, the
other properties involved (polynomiality and modularity) are in effect
the same.

\subsection{Input representation $F(\Pi)$}
\label{sec:transinput}

As input for our meta-interpreter $\Pi_{meta}$, which will be
introduced in the next subsection, we choose the representation
$F(\Pi)$ of the propositional program $\Pi$ defined below. We assume
that each rule $r$ has a unique name $n(r)$; for convenience, we
identify $r$ with $n(r)$.

\begin{definition}
Let $\Pi$ be any ground (propositional) HDLP. The set $F(\Pi)$ consists of the facts
{
\begin{tabbing}
XXX\=\kill
\> \texttt{lit(h,$l,r$).\ \ atom($l$,$|l|$).} \quad \= for each literal $l$ $\in \head{r}$,\+\\
\texttt{lit(p,$l,r$).} \>for each literal $l \in \pbody{r}$, \\
\texttt{lit(n,$l,r$).} \>for each literal $l \in \nbody{r}$,\-
\end{tabbing}
}
\noindent for every rule $r \in \Pi$.
\end{definition}

While the facts for predicate \texttt{lit} obviously
encode the rules of $\Pi$, the facts for predicate \texttt{atom}
indicate whether a literal is classically positive or negative.
We only need this information for head literals; this will be further
explained below.

\subsection{Meta-Interpreter $\Pi_{meta}$}
\label{sec:transmeta}

We construct our meta-interpreter program $\Pi_{meta}$, which in
essence is a positive disjunctive program, in a sequence of several
steps. They center around checking whether a guess for an answer set
$S \subseteq Lit(\Pi)$, encoded by a predicate \texttt{inS($\cdot$)},
is an answer set of $\Pi$ by testing the criteria of
Theorem~\ref{theo:hedlp}. The steps of the transformation cast the various
conditions there into rules of $\Pi_{meta}$, and also provide
auxiliary machinery which is needed for this aim.

\newcounter{bctr}
\newcommand{\incb}{\addtocounter{bctr}{1}\thebctr}

\paragraph{Step 1} We add the following preprocessing rules:

{\small\tt
\begin{tabbing}
\incb: \ \ \ \=rule(L,R) \derives lit(h,L,R), not lit(p,L,R), not lit(n,L,R).\\[0.33ex]
\incb: \>ruleBefore(L,R) \derives rule(L,R), rule(L,R1), R1 < R.\\[0.33ex]
\incb: \>ruleAfter(L,R) \derives rule(L,R), rule(L,R1), R < R1.\\[0.33ex]
\incb: \>ruleBetween(L,R1,R2) \derives \=rule(L,R1), rule(L,R2), rule(L,R3),\\ \>\>R1 < R3, R3 < R2.\\[1.0ex]
\incb: \>firstRule(L,R) \derives rule(L,R), \dneg{ruleBefore(L,R)}.\\[0.33ex]
\incb: \>lastRule(L,R) \derives rule(L,R), \dneg{ruleAfter(L,R)}.\\[0.33ex]
\incb: \>nextRule(L,R1,R2) \derives \=rule(L,R1), rule(L,R2), R1 < R2,\\ \>\>\dneg{ruleBetween(L,R1,R2)}.\\[1.0ex]
\incb: \>before(HPN,L,R) \derives lit(HPN,L,R), lit(HPN,L1,R), L1 < L.\\[0.33ex]
\incb: \>after(HPN,L,R) \derives lit(HPN,L,R), lit(HPN,L1,R), L < L1.\\[0.33ex]
\incb: \>between(HPN,L,L2,R) \derives \=lit(HPN,L,R), lit(HPN,L1,R),\\ \>\>lit(HPN,L2,R), L<L1, L1<L2.\\[0.33ex]
\incb: \>next(HPN,L,L1,R) \derives \=lit(HPN,L,R), lit(HPN,L1,R), L < L1,\\ \>\>\dneg{between(HPN,L,L1,R)}.\\[0.33ex]
\incb: \>first(HPN,L,R) \derives lit(HPN,L,R), \dneg{before(HPN,L,R)}.\\[0.33ex]
\incb: \>last(HPN,L,R) \derives lit(HPN,L,R), \dneg{after(HPN,L,R)}.\\[1.0ex]
\incb:  \>hlit(L) \derives rule(L,R).
\end{tabbing}
}


Lines 1 to 7 fix an enumeration of the rules in $\Pi$ from which
a literal $l$ may be derived, assuming a given order \texttt{<} on rule
names (e.g.\ in \dlv, built-in lexicographic order; \texttt{<} can also
be easily generated using guessing rules). Note that under answer set semantics, we
need only to consider rules where the literal $l$ to prove does not occur in the
body.

\noindent
Lines 8 to 13 fix enumerations of $\head{r}$, $\pbody{r}$ and $\nbody{r}$
for each rule. The final line 14 collects all literals that can be
derived from rule heads. Note that the rules on lines 1-14 plus
$F(\Pi)$ form a stratified program,
which has a single answer set, cf.\ \cite{przy-89b,przy-91}.

\paragraph{Step 2}
Next, we add rules which ``guess'' a candidate answer set
$S \subseteq Lit(\Pi)$ and a total ordering
\texttt{phi} on $S$ corresponding with the function
$\phi$ in condition $2$ of Theorem~\ref{theo:hedlp}.
We will explain this correspondence in more detail below (cf.\ proof of
Theorem~\ref{theo:corr}).


{\small\tt
\begin{tabbing}
\incb: \ \ \ \=inS(L) \vel\ ninS(L) \derives hlit(L).\\[0.33ex]
\incb: \>ninS(L) \derives lit($pn$,L,R), \dneg{hlit(L)}.
\ \ \ \ \ \ \ \ \ \textrm{for each $pn \in \{$\texttt{p,n}$\}$}\\[0.33ex]
\incb: \ \ \ notok \derives inS(L), inS(NL), L\kuneq NL, atom(L,A), atom(NL,A).\\[0.33ex]
\incb: \>phi(L,L1) \vel\ phi(L1,L) \derives inS(L), inS(L1), L < L1.\\[0.33ex]
\incb: \>phi(L,L2) :- phi(L,L1),phi(L1,L2).

\end{tabbing}
}
Line~15 focuses the guess of $S$ to literals occurring in some relevant rule
head in $\Pi$; only these can belong to an answer set $S$, but no others (line~16).
Line~17 then checks whether $S$ is consistent, deriving a new distinct atom
\texttt{notok} otherwise.
 Line~18 guesses a strict total order \texttt{phi} on \texttt{inS} where line~19 guarantees
transitivity; note that minimality of answer sets prevents that
\texttt{phi} is cyclic, i.e., that \texttt{phi(L,L)} holds.

\medskip
 In the subsequent steps, we will check whether $S$ and \texttt{phi}
violate the conditions of Theorem~\ref{theo:hedlp} by deriving the
distinct atom \texttt{notok} (considered in Step~5 below) in case,
indicating that $S$ is not an answer set or \texttt{phi} does not represent a
proper function $\phi$.

\paragraph{Step 3} Corresponding to condition $1$ in
Theorem~\ref{theo:hedlp}, \texttt{notok} is derived whenever there
is an unsatisfied rule by the following program part:


{\small\tt
\begin{tabbing}
\incb:\ \ \ \=allInSUpto(p,Min,R) \derives inS(Min), first(p,Min,R).\\[0.33ex]
\incb: \>allInSUpto(p,L1,R) \derives \=inS(L1), allInSUpto(p,L,R), next(p,L,L1,R).\\[0.33ex]
\incb: \>allInS(p,R) \derives allInSUpto(p,Max,R),last(p,Max,R).
\end{tabbing}
\hspace{-0.2ex}$\left.\mbox{\begin{minipage}{0.9\textwidth}
\begin{tabbing}
\incb:\ \ \ \=allNinSUpto($hn$,Min,R) \derives ninS(Min), first($hn$,Min,R).\\[0.33ex]
\incb: \>allNinSUpto($hn$,L1,R) \derives \=ninS(L1), allNinSUpto($hn$,L,R),\\ \>\>next($hn$,L,L1,R).\\[0.33ex]
\incb: \>allNinS($hn$,R) \derives allNinSUpto($hn$,Max,R), last($hn$,Max,R).
\end{tabbing}
\end{minipage}
\hspace{-1ex}}\right\}$
\mbox{\begin{minipage}{0.15\textwidth}
\textrm{for each\\ $hn \in \{$\texttt{h,n}$\}$}
\end{minipage}
}
\vspace{-1.5ex}
\begin{tabbing}
\incb:\ \ \ \=hasHead(R) \derives lit(h,L,R).\\[0.33ex]
\incb: \> hasPBody(R) \derives lit(p,L,R).\\[0.33ex]
\incb: \> hasNBody(R) \derives lit(n,L,R).\\[0.33ex]
\incb: \> allNinS(h,R) \derives lit(HPN,L,R), \dneg{hasHead(R)}.\\[0.33ex]
\incb: \> allInS(p,R) \derives lit(HPN,L,R), \dneg{hasPBody(R)}.\\[0.33ex]
\incb: \> allNinS(n,R) \derives lit(HPN,L,R), \dneg{hasNBody(R)}.\\[1.0ex]
\incb: \>notok \derives allNinS(h,R), allInS(p,R), allNinS(n,R), lit(HPN,L,R).
\end{tabbing}
}

These rules compute by iteration over $\pbody{r}$
(resp.\ $\head{r}$, $\nbody{r}$) for each rule $r$,
whether for all positive body (resp.\ head and
default negated body) literals in rule $r$ \texttt{inS} holds
(resp.\ \texttt{ninS} holds) (lines 20 to 25). Here, empty heads
(resp. bodies) are interpreted as unsatisfied (resp.\ satisfied),
cf.\ lines 26 to 31. The final rule 32 fires exactly if one of the
original rules from $\Pi$ is unsatisfied.

\paragraph{Step 4} We derive \texttt{notok} whenever
there is a literal $l\in S$ which is not provable by any rule $r$
with respect to \texttt{phi}. This corresponds to checking condition $2$ from
Theorem~\ref{theo:hedlp}.


{\small\tt
\begin{tabbing}
\incb:\ \ \ \=failsToProve(L,R) \derives rule(L,R), lit(p,L1,R), ninS(L1).\\[0.33ex]
\incb: \>failsToProve(L,R) \derives rule(L,R), lit(n,L1,R), inS(L1).\\[0.33ex]
\incb: \>failsToProve(L,R) \derives rule(L,R), rule(L1,R), inS(L1), L1\kuneq L, inS(L).\\[0.33ex]
\incb: \>failsToProve(L,R) \derives rule(L,R), lit(p,L1,R), phi(L1,L).\\[1.0ex]
\incb: \>allFailUpto(L,R) \derives failsToProve(L,R), firstRule(L,R).\\[0.33ex]
\incb: \>allFailUpto(L,R1) \derives \=failsToProve(L,R1), allFailUpto(L,R),\\ \>\>nextRule(L,R,R1).\\[0.33ex]
\incb: \>notok \derives allFailUpto(L,R), lastRule(L,R), inS(L).
\end{tabbing}
}


Lines 33 and 34 check whether condition $2.(a)$ or $(b)$ are violated,
i.e.\ some rule can only prove a literal if its body is satisfied.
Condition $2.(d)$ is checked in line~35, i.e. $r$ fails to prove $l$
if there is some $l'\neq l$  such that $l' \in \head{r}\cap S$.
Violations of condition $2.(e)$ are checked in line 36.
Finally, lines 37 to 39 derive \texttt{notok} if  all rules fail to
prove some literal $l\in S$. This is checked by iterating
over all rules with $l\in \head{r}$ using the order from Step~1.
Thus, condition $2.(c)$ is implicitly checked by this iteration.

\paragraph{Step 5} Whenever \texttt{notok} is derived, indicating a
wrong guess, then we
apply a saturation technique as in \cite{eite-etal-97f,eite-etal-2000c,leon-etal-2001} to some other predicates, such that
a canonical set $\Omega$ results. This set turns out to be an answer
set iff no guess for $S$ and $\phi$ works out, i.e., $\Pi$ has no answer set.
In particular, we saturate the predicates \texttt{inS}, \texttt{ninS}, and \texttt{phi} by the
following rules:


{\small\tt
\begin{tabbing}
\incb:\ \ \ \= phi(L,L1)\ \=\derives notok, hlit(L), hlit(L1).\\[0.33ex]
\incb: \>inS(L)     \derives notok, hlit(L).\\[0.33ex]
\incb: \>ninS(L)    \derives notok, hlit(L).
\end{tabbing}
}

Intuitively, by these rules, any answer set containing \texttt{notok}
is ``blown up'' to an answer set $\Omega$ containing all
possible guesses for \texttt{inS}, \texttt{ninS}, and
\texttt{phi}.

\begin{definition} The program $\Pi_{meta}$ consists of the
rule 1--42 from above.
\end{definition}

We then can formally define our transformation $\tr{\Pi}$ as follows.

\begin{definition}
Given any ground HDLP $\Pi$, its transformation $\tr{\Pi}$ is given by
the DLP $\tr{\Pi}= F(\Pi) \cup \Pi_{meta}$.
\end{definition}

Examples of $\tr{\Pi}$ will be provided in Section~\ref{sec:applications}.

\subsection{Properties of $\tr{\Pi}$}

We now show that $\tr{\Pi}$ satisfies indeed the
properties {\bf T0} -- {\bf T4} from the beginning of this section.

As for {\bf T0}, we note the following proposition, which is not
difficult to establish.

\begin{proposition}
\label{prop:poly}
Given $\Pi$, the transformation $\tr{\Pi}$ and its ground
instantiation are both computable in logarithmic workspace
(and thus in polynomial time).
\end{proposition}

\begin{proof} The input representation $F(\Pi)$ is easily generated
in a linear scan of $\Pi$, using the rule numbers as names, for which
a counter (representable in logspace) is sufficient. The
meta-interpreter part $\Pi_{meta}$ is fixed anyway. A naive grounding
of $\tr{\Pi}$ can be constructed by instantiating each rule $r$ from
$\Pi_{meta}$ with constants from $\Pi$ and rule ids in all possible
ways; for each variable \texttt{X} in $r$, all constants of $\Pi$ can be
systematically considered, using counters to mark the start and end
position in $\Pi$ (viewed as a string), and the rule ids by a rule
number counter. A constant number of such counters is
sufficient. Thus, the grounding of $\tr{\Pi}$ is constructible in
logarithmic work space. Notice that intelligent, efficient grounding
methods such as those used in \dlv \cite{leon-etal-2002-dlv} usually
generate a smaller ground program than this naive ground instantiation.
\end{proof}

Clearly, $\tr{\Pi}$ satisfies property {\bf T3}, and as easily checked, $\tr{\Pi}$ is modular.
Moreover, strong negation does not occur in $\trP(\Pi)$ and weak negation only stratified. The latter is
not applied to literals depending on disjunction; it thus occurs only
in the deterministic part of $\trP(\Pi)$, which means {\bf T4} holds.

To establish {\bf T1} and {\bf T2}, we define the literal set $\Omega$ as follows:
\begin{definition}
Let $\Pi_{meta}^i$ be the set of rules  in $\Pi_{meta}$ established in
Step~$i \in \{1,\ldots,5\}$. For any program $\Pi$, let
$\Pi_{\Omega} = F(\Pi) \cup \bigcup_{i\in \{1,3,4,5\}}\Pi_{meta}^i \cup \set{\texttt{notok}.}$.
Then, $\Omega$ is defined as the answer set of $\Pi_{\Omega}$.
\end{definition}

\begin{lemma}
$\Omega$ is well-defined and uniquely determined by $\Pi$.
\end{lemma}
\begin{proof}(Sketch) This follows immediately from the fact that $\Pi_{\Omega}$
is a (locally) stratified normal logic program without $\tneg{}$ and constraints,
which as well-known has a single
answer set.
\end{proof}


\begin{theorem}
\label{theo:corr}
For a given HDLP $\Pi$ the following holds for $\tr{\Pi}$:
\begin{enumerate}
\item $\tr{\Pi}$ always has some answer set, and $S' \subseteq
        \Omega$ for every answer set $S'$ of $\tr{\Pi}$.
\item $S$ is an answer set of $\Pi$ $\Leftrightarrow$
there exists an answer set $S'$ of $\tr{\Pi}$ such that
$S=\set{ l \mid \mathtt{inS(\mbox{$l$})} \in S'}$ and $\texttt{notok} \not\in S'$.
\item $\Pi$ has no answer set $\Leftrightarrow$ $\tr{\Pi}$ has the
unique answer set $\Omega$.
\end{enumerate}
\end{theorem}

\begin{proof} $1.$\ The first part follows immediately from the fact that
$\tr{\Pi}$ has no constraints, no strong negation, and weak negation is stratified; this
guarantees the existence of at least one answer set $S$ of
$\tr{\Pi}$~\cite{przy-91}. Moreover, $S'\subseteq\Omega$ must hold for
every answer set: after removing $\{\texttt{notok.}\}$ from $\Pi_{\Omega}$
and adding $\Pi_{meta}^{2}$, we obtain $\tr{\Pi}$. Note that any rule in
$\Pi_{meta}^{2}$ fires with respect to $S'$ only if all literals in its head are
in $\Omega$, and \texttt{inS}, \texttt{ninS}, and \texttt{phi} are
elsewhere not referenced recursively through negation or
disjunction. Therefore, increasing $S'$ locally to the value of $\Omega$ on
\texttt{inS}, \texttt{ninS}, \texttt{phi}, and \texttt{notok}, and
closing off thus increases it globally to $\Omega$, which means $S'
\subseteq \Omega$.

\noindent $2.$\ ($\Rightarrow$)\  Assume that $S$ is an answer set of $\Pi$. Clearly,
then $S$ is a consistent set of literals which has a corresponding
set $S''= \set{ \texttt{inS}(l) \mid l \in S}$ $\cup$
$\set{\texttt{ninS}(l) \mid l \in Lit(\Pi)\setminus S}$ being one possible guess by the rules
in lines 15 to 17 of  $\Pi_{meta}$.
Let now $\phi: Lit(\Pi) \rightarrow \Nat$ be the function
from Theorem~\ref{theo:hedlp} for answer set $S$:
Without loss of generality, we may assume two restrictions
on this function $\phi$:
\begin{itemize}
\item $\phi(l) = 0$ for all $l\in Lit(\Pi)\setminus S$ and $\phi(l) >
0$ for all $l\in S$.
\item $\phi(l)\neq\phi(l')$ for all $l,l' \in S$.
\end{itemize}
Then, the function $\phi$ can be mapped to a total order over $S$
\texttt{phi} such that
$$
\mathtt{phi(\mbox{$l,l'$})} \Leftrightarrow \phi(l) > \phi(l') > 0.
$$
This relation \texttt{phi} fixes exactly one possible guess by the
lines 18 and 19 of $\Pi_{meta}$.

Note that it is sufficient to define \texttt{phi} only over literals
in $S$: Violations of condition $2.(e)$  have only to be checked
for rules with $\pbody{r} \subseteq S$, as otherwise condition $2.(a)$
already fails. Obviously, condition 2.$(e)$ of Theorem~\ref{theo:hedlp}
is violated with respect to $\phi$ iff (a) \texttt{phi(Y,X)} holds for some
\texttt{X} in the head of a rule with \texttt{Y} in its positive
body or (b) if \texttt{X} itself occurs in its positive body.
While (a) is checked in lines 36, (b) is implicit by definition
of predicate \texttt{rule} (line 1) which says that a literal can
not prove itself.

Given $S''$ and \texttt{phi} from above, we can now verify by our
assumption that $S$ is an answer set and by the conditions of
Theorem~\ref{theo:hedlp} that (a) \texttt{notok} can never be derived in
$\tr{\Pi}$ and (b) $S''$ and \texttt{phi} uniquely determine an answer
set $S'$ of $\tr{\Pi}$
of the form we want to prove. This can be argued
by construction of Steps 3 and 4 of $\tr{\Pi}$, where \texttt{notok}
will only be derived if some rule is unsatisfied (Step 3) or there is
a literal in $S$ (i.e.\ $S''$) which fails to be proved by all other
rules (Step 4).

($\Leftarrow$)\
Assume that $S'$ is an answer set of \tr{\Pi} not containing
\texttt{notok}. Then by the guess of \texttt{phi} in Step 5
a function $\phi: Lit(\Pi) \rightarrow \Nat$ can be constructed
by the implied total order of \texttt{phi} as follows:
We number all literals $l \in S=\set{l \mid \mathtt{inS(\mbox{$l$})} \in S'}$
according to that order from $1$ to $|S|$ and fix $\phi(l)=0$ for
all other literals.
Again, by construction of Steps 3 to 5 and the assumption
that $\texttt{notok} \not\in S'$, we can see that $S$ and the
function $\phi$ constructed fulfill all the
conditions of Theorem~\ref{theo:hedlp}; in particular, line~17
guarantees consistency. Hence $S$ is an answer set of $\Pi$.

\noindent $3.$\ ($\Leftarrow$)\  Assume that $\Pi$ has an answer set. Then, by the
already proved Part~2 of the Theorem,
we know that there exists an answer set $S'$ of $\tr{\Pi}$ such that \texttt{notok} $\not\in S'$.
By minimality of answer sets, $\Omega$ can not be an answer
set of $\tr{\Pi}$.

($\Rightarrow$)\  By Part~1 of Theorem~\ref{theo:corr}, we know
that $\tr{\Pi}$ always has an answer set $S' \subseteq
        \Omega$. Assume that there is an answer set
        $S'\subsetneqq\Omega$.
We distinguish 2 cases: (a) \texttt{notok} $\not\in S'$ and
(b) \texttt{notok} $\in S'$. In case (a), proving Part~2 of this
proposition, we have already shown that $\Pi$ has an answer set; this
is a contradiction. On the other hand, in case (b)
the final ``saturation'' rules in Step 5 ``blow up'' any answer set
containing \texttt{notok} to $\Omega$, which contradicts the assumption
$S'\subsetneqq\Omega$.
\end{proof}

As noticed above, the transformation $\tr{\Pi}$ uses weak negation
only stratified and in a deterministic part of the program; we can
easily eliminate it by computing in the
transformation the complement of each predicate accessed through
$\naf$ and providing it in $F(\Pi)$ as facts; we then obtain a positive
program. (The built-in predicates \texttt{$<$} and \texttt{$!=$}
can be eliminated similarly if desired.) However, such a modified
transformation is not modular. As shown next, this is not
incidental.

\begin{proposition}
\label{prop:non-modular}
There is no modular transformation $\trP'(\Pi)$ from HDLPs to DLPs
(i.e.\ such that
$\trP'(\Pi) = \bigcup_{r \in \Pi} \trP'(r)$),  satisfying {\bf T1}
such that $\trP'(\Pi)$ is a
positive program.
\end{proposition}

\begin{proof}
Assuming such a transformation exists, we derive a contradiction. Let
$\Pi_1 = \{ \texttt{ a \derives \naf b.} \}$ and $\Pi_2 = \Pi_1 \cup
\{ \texttt{b.} \}$. Then, $\trP'(\Pi_2)$ has some answer set
$S_2$. Since $\trP'(\cdot)$ is modular, $\trP'(\Pi_1) \subseteq
\trP'(\Pi_2)$ holds and thus $S_2$ satisfies each rule in
$\trP'(\Pi_1)$. Since $\trP'(\Pi_1)$ is a positive program, $S_2$
contains some answer set $S_1$ of $\trP'{\Pi_1}$. By {\bf
  T1}, we have that \texttt{inS(a)} $\in S_1$ must hold, and hence  \texttt{inS(a)}
$\in S_2$. By {\bf T1} again, it follows that $\Pi_2$ has an answer set
$S$ such that $\texttt{a} \in S$. But the single answer set of $\Pi_2$ is
$\{ \texttt{b} \}$, which is a contradiction.
\end{proof}

We remark that Prop.~\ref{prop:non-modular} remains true if {\bf T1}
is generalized such that the answer set $S$ of $\Pi$ corresponding to
$S'$ is given by $S=\{l \mid S'\models \Psi(l)\}$, where $\Psi(x)$ is
a monotone query (e.g., computed by a normal positive program without
constraints).  Moreover, if a successor predicate \texttt{next(X,Y)}
and predicates \texttt{first(X)} and \texttt{last(X)} for the
constants are available, given that the universe is finite by
the constants in $\Pi$ and rule names, then computing the negation of the
non-input predicates accessed through $\naf$ is feasible by a positive
normal program, since such programs capture polynomial time
computability by well-known results on the expressive power of
Datalog~\cite{papa-85}; thus, negation of input predicates in
$F(\Pi)$ is sufficient in this case.

\section{Modifications towards Optimization}
\label{sec:optimizations}

The meta-interpreter $\Pi_{meta}$ from above can be modified in several
respects. We discuss in this section some modifications which, though not
necessarily reducing the size of the ground instantiation, intuitively
prune the search of an answer set solver applied to $\tr{\Pi}$.

\subsection{Giving up modularity (OPT$_{mod}$)}

If we sacrifice modularity and allow that $\Pi_{meta}$
partly depends on the input, then we can circumvent
the iterations in Step~3 and in part of Step~1.
Intuitively, instead of
iterating over the heads and bodies of all rules in order to determine
whether these rules are satisfied, we add a single rule in $\tr{\Pi}$ for
each 
rule $r$ in $\Pi$ firing \inlinek{notok} whenever $r$
is unsatisfied. We therefore replace the rules from Step~3 by
\begin{equation}
\label{stmt:rulesatisfied}
\begin{split}
\mathtt{notok\ \derives\ }&\mathtt{ninS(\mbox{$h_1$}),\ \ldots,\ ninS(\mbox{$h_l$}),\
inS(\mbox{$b_1$}),\ \ldots,\ inS(\mbox{$b_m$}),}\\
& \mathtt{ninS(\mbox{$b_{m+1}$}),\ \ldots\, ninS(\mbox{$b_n$}).}
\end{split}
\end{equation}
for each rule $r$ in $\Pi$ of form (\ref{stmt:lprule}). These rules
can be efficiently generated in parallel to $F(\Pi)$. Lines 8 to 13 of
Step 1 then become unnecessary and can be dropped.

\medskip
We can even refine this further. For every normal rule $r \in \Pi$ with non-empty head, i.e.\
$\head{r}=\set{h}$, which has a satisfied body, we can force the guess
of $h$: we replace
(\ref{stmt:rulesatisfied}) by
\begin{equation}
\label{stmt:forcenormal}
\mathtt{inS(\mbox{$h$}) \derives\ inS(\mbox{$b_1$}),\ \ldots,\ inS(\mbox{$b_m$}), ninS(\mbox{$b_{m+1}$}),\ \ldots\ ninS(\mbox{$b_n$}).}
\end{equation}


In this context, since constraints only serve to ``discard'' unwanted
models but cannot prove any literal, we can ignore them during input
generation $F(\Pi)$.
Note that dropping input representation  $\mathtt{lit(n,\mbox{$l$},\mbox{$c$}).}$
for literals only occurring in the negative body of constraints
but nowhere else in $\Pi$ requires some care.
Such $l$ can be removed by simple preprocessing, though, by removing
all $l \in \nbody{c}$ which do not occur in any rule head in $\Pi$.
On the other hand, all literals $l\in \nbody{c}$ which appear in some
other (non-constraint) rule $r$ are not critical, since facts
\texttt{lit($hpn$,$l$,$r$).} ($hpn \in \{\texttt{h,p,n}\}$) from this
other rule will ensure that either line~15 or line~16 in $\Pi_{meta}$
is applicable and therefore, either \texttt{inS($l$)} or
\texttt{ninS($l$)} will be derived. Thus, after elimination of
critical literals in constraints beforehand,
we can safely drop the factual representation
of constraints completely (including \texttt{lit(n,$l,c$).} for the
remaining negative literals).

\subsection{Restricting to potentially applicable rules (OPT$_{pa}$)}

We only need to consider literals in heads of {\em potentially
applicable} rules. These are all rules with empty bodies, and rules where any
positive body literal -- recursively -- is the head of
another potentially applicable rule. This suggests the following definition:
\begin{definition}
\label{def:poss_app}
A set $R$ of ground rules is \emph{potentially applicable}, if there
exists an enumeration $\tuple{r_i}_{i \in I}$ of $R$, where $I$ is a prefix of $\Nat$ resp.\ $I{=}\Nat$, such that
$\pbody{r_i} \subseteq \bigcup_{j<i} \head{r_j}$.
\end{definition}

The following proposition is then not difficult to establish.

\begin{proposition}
Let  $\Pi$ be any ground HDLP. Then there exists a unique maximal
set $R^* \subseteq \Pi$ of potentially applicable rules, denoted by
$\mathrm{PA}(\Pi)$.
\end{proposition}

\begin{proof} Indeed, suppose  $\tuple{r_i}_{i \in I}$ and
$\tuple{r'_i}_{i \in I'}$ are enumerations witnessing that rule sets
    $R$ and $R'$ such that $R,R'\subseteq \Pi$
are potentially applicable. Then their union $R\cup R'$ is potentially
applicable, witnessed by the enumeration obtained from the alternating
enumeration $r_0,r'_0,r_1,r'_1$,\ldots whose suffix are the
rules from the larger set of $R$ and $R'$ if they have different
cardinalities, from which duplicate rules are removed (i.e., remove
any rule $r'_j$ if $r'_j=r_i$, for some $i\leq j$, and remove any rule
$r_j$ if $r'_i=r_j$ and for some $i<j$). It follows that a unique
largest set $R^* \subseteq \Pi$ of potentially applicable rules
exists.
\end{proof}

\noindent
The set $\mathrm{PA}(\Pi)$ can be computed by adding a rule:
{\tt
\begin{tabbing}
\ \  \=pa($r$) :- lit(h,$b_1$,R$_1$), pa(R$_1$), \ldots, lit(h,$b_m$,R$_m$), pa(R$_m$).
\end{tabbing}
}
\noindent for any rule $r$ of the form~(\ref{stmt:lprule}) in $\Pi$. In particular,
if $m=0$ we simply add the fact \texttt{\small pa($r$).}
Finally, we change line~1 in $\Pi_{meta}$ to:
{\tt
\begin{tabbing}
\ \  \=rule(L,R) \derives lit(h,L,R), not lit(p,L,R),  not lit(n,L,R), pa(R).
\end{tabbing}
}
\noindent such that only ``interesting'' rules are considered.

We note, however, that computing \texttt{pa($\cdot$)} incurs some
cost: Informally, a profit of optimization \textbf{OPT$_{pa}$} might
only be expected in domains where $\Pi_{check}$ contains a a
reasonable number of rules which positively depend on each other and
might on the other hand likely be ``switched off'' by particular
guesses in $\Pi_{guess}$.

\subsection{Optimizing the order guess (OPT$_{dep}$)}

We only need to guess and check the order $\phi$ for literals
$L$, $L'$ if they allow for cyclic dependency, i.e., they appear in the
heads of rules within the same strongly connected component of the program
with respect to $S$.%
\footnote{Similarly, in~\cite{bene-dech-94}
$\phi: Lit(\Pi) \rightarrow \{1, \ldots, r\}$
is only defined for a range $r$ bound by
the longest acyclic path in any strongly connected component of the
program.}  These dependencies with respect to $S$ are easily computed:
{\tt
\begin{tabbing}
\ \  \=dep(L,L1) \derives lit(h,L,R),lit(p,L1,R),inS(L),inS(L1).\\[0.33ex]
\> dep(L,L2) \derives lit(h,L,R),lit(p,L1,R),dep(L1,L2),inS(L).\\[0.33ex]
\> cyclic \derives dep(L,L1),dep(L1,L).
\end{tabbing}
}
\noindent
The guessing rules for $\phi$ (line 18 and 19) are then be replaced by:
{\tt
\begin{tabbing}
\ \  \=phi(L,L1) \vel\ phi(L,L1) \derives dep(L,L1), dep(L1,L), L < L1,cyclic.\\[0.33ex]
\> phi(L,L2) :- phi(L,L1),phi(L1,L2), cyclic.
\end{tabbing}
}
\noindent Moreover, we add the new atom \texttt{cyclic} also to the body of
any other rule where \texttt{phi} appears (lines 36,40) to
check \texttt{phi} only in case $\Pi$ has \emph{any} cyclic
dependencies with respect to $S$.

In the following, we will denote the transformation obtained by the
optimizations from this section as $\trOpt{\Pi}$ while we refer to
$\tr{\Pi}$ for the original transformation.

\section{Integrating Guess and $\coNP$ Check Programs}
\label{sec:integration}

In this section, we show how our transformation $\trP$ (resp.\
$\trP_{Opt}$) from above can be used to automatically combine a HDLP
$\Pi_{guess}$ which guesses in its answer sets solutions of a problem,
and a HDLP $\Pi_{check}$ which encodes a $\coNP$-check of the solution
property, into a single DLP $\Pi_{solve}$ of the form $\Pi_{solve} =
\Pi_{guess} \cup \Pi'_{check}$.

We assume that the set $Lit(\Pi_{guess})$ is a Splitting Set
\cite{lifs-turn-94} for $\Pi_{guess} \cup \Pi_{check}$, i.e.\ no head
literal from $\Pi_{check}$ occurs in $\Pi_{guess}$. This can be easily
achieved by introducing new predicate names, e.g., $\texttt{p}'$ for a
predicate $\texttt{p}$, and adding a rule $\texttt{p}'(t) \derives
\texttt{p}(t)$ in case there is an overlap.

Each rule $r$ in $\Pi_{check}$ is of the form
\begin{equation}\label{eqn:int}
\begin{split}
h_1 \vel\ \cdots\ \vel\ h_l\ \derives\ & bc_1,\ \ldots,\ bc_m,\
\dneg{bc_{m+1}},\ \ldots,\ \dneg{bc_n}\\
& bg_1,\ \ldots,\ bg_p,\
\dneg{bg_{p+1}},\ \ldots,\ \dneg{bg_q}.
\end{split}
\end{equation}
where the $bg_i$ are the body literals
defined in $\Pi_{guess}$. We write ${\sf body}_{guess}(r)$ for $bg_1,\
\ldots,\ bg_p,\ \dneg{bg_{p+1}},\ \ldots,\ \dneg{bg_q}$.
We now define a new check program as follows.

\begin{definition}
For any ground program $\Pi_{check}$ as above, the program $\Pi'_{check}$ contains the following rules and constraints:
\begin{description}
\item[] (i) The facts $F(\Pi_{check})$ in a conditional version: For
each rule $r\in\Pi_{check}$ of form (\ref{eqn:int}), the rules

{
\begin{tabbing}
 \quad \ \ \ \ \=\texttt{lit(p,$bg_i,r$)\derives}\=\kill
\>\> \texttt{lit(h,$l,r$)\derives}$\!\!$ \'$\!\!\!\!${\sf body}$_{guess}(r).$ \quad \texttt{atom($l$,$|l|$).} \` for each $l \in \head{r}$;\\
\>\>\texttt{lit(p,$bc_i,r$)\derives}$\!\!$ \'$\!\!\!\!${\sf body}$_{guess}(r).$ \` for each
$i \in \{1,\ldots, m\}$;\\
\>\>\texttt{lit(n,$bc_j,r$)\derives}$\!\!$ \'$\!\!\!\!${\sf body}$_{guess}(r).$ \` for each $j \in \{m+1,\ldots, n\}$;
  \end{tabbing}
}

\item[] (ii) each rule in $\Pi_{meta}{\,=\,}\tr{\Pi_{check}}\!\setminus F(\Pi_{check})$
(resp.\ in $\trOpt{\Pi_{check}}\!\setminus\!F(\Pi_{check})$, where
${\sf body}_{guess}(r)$ must be added to the bodies of the rules
(\ref{stmt:rulesatisfied}) and (\ref{stmt:forcenormal}));
\item[] (iii) a constraint
{
\tt
\begin{tabbing}
 \ \ \ \=\derives\ \dneg{notok}.
\end{tabbing}
}
\noindent It eliminates any answer set $S$ such that
$\Pi_{check} \cup S$ has an answer set.
\end{description}
\end{definition}

The union of $\Pi_{guess}$ and $\Pi_{check}'$ then amounts to the
desired integrated encoding $\Pi_{solve}$, which is
expressed by the following result.

\begin{theorem}
\label{theo:integrate}
Given separate guess and check programs $\Pi_{guess}$ and $\Pi_{check}$, the answer sets of
\[
\Pi_{solve}=\Pi_{guess}\cup\Pi_{check}',
\] denoted  $S_{solve}$, are in  1-1 correspondence
with the answer sets $S$ of $\Pi_{guess}$ such that
$\Pi_{check}\!\cup\!S$ has no answer set.
\end{theorem}

\begin{proof}  This result can be derived from Theorem~\ref{theo:corr} and the
Splitting Set Theorem for logic programs under answer set semantics \cite{lifs-turn-94}. We consider the proof for
the original transformation $\tr{\cdot}$; the proof for the optimized
transformation $\trOpt{\cdot}$ is similar (with suitable extensions in places). In what follows, for
any program $Q$ and any consistent literal set $S$, we let $Q[S]$ denote the program
obtained from $Q$ by eliminating every rule $r$ such that
${\sf body}_{guess}(r)$ is false in $S$,  and by removing
${\sf body}_{guess}(r)$
from the remaining rules. Notice that $\Pi_{check}\cup S$ and
$\Pi_{check}[S]\cup S$ have the same answer sets.

We can rewrite $\Pi_{solve}$ as
$$
\Pi_{solve} = \Pi_{guess} \cup F'(\Pi_{check}) \cup \Pi_{meta} \cup
\{\,\derives\ \dneg{\texttt{notok}}.\,\}
$$
\noindent where $F'(\Pi_{check})$ denotes the modified factual
representation for $\Pi_{check}$, given in item $1.$ of the definition
of $\Pi'_{check}$. By hypothesis on $\Pi_{guess}\cup \Pi_{check}$, the
set $Lit(\Pi_{guess})$ is a splitting set for $\Pi_{solve}$.
Hence, as easily seen also $Lit(\Pi_{guess} \cup F'(\Pi_{check}))$ is
a splitting set for $\Pi_{solve}$, and $Lit(\Pi_{guess})$ is also a
splitting set for $\Pi_{guess} \cup F'(\Pi_{check})$. Moreover, each
answer set $S$ of $\Pi_{guess}$ is in 1-1 correspondence with
an answer set $S'$ of $\Pi_{guess} \cup F'(\Pi_{check})$. Then $S'
\setminus S = F(\Pi_{check}[S]) \cup A_S$, such that $F(\Pi_{check}[S])$ is
the factual representation of $\Pi_{check}[S]$ in the transformation
$\tr{\Pi_{check}[S]}$ and $A_S = \set{\texttt{atom}(l,|l|). \mid l \in
\head{\Pi_{check}}\setminus \head{\Pi_{check}[S]}}$\footnote{Here,
for any program $\Pi$, we write $\head{\Pi}= \bigcup_{r \in \Pi} \head{r}$.}
is an additional set of facts emerging from $F'(\Pi_{check})$, since we added
facts \texttt{atom}$(l,|l|)$. for all head literals of
$r \in \Pi_{check}$, not only for those $r$ where
${\sf body}_{guess}(r)$ was satisfied.

Now let $S_{solve}$ be any (consistent) answer set of $\Pi_{solve}$.
From the Splitting Set Theorem \cite{lifs-turn-94}, we can conclude
that $S_{solve}$ can be written as $S_{solve} =
S \cup S_{check} \cup A_S$ where $S$ and $S_{check} \cup A_S$ are
disjoint,
$S$ is an answer set of $\Pi_{guess}$, and $S_{check} \cup A_S$
is an answer set of the program $\Pi'_S = (\Pi_{solve} \setminus
\Pi_{guess})[S]$.
Since $F'(\Pi_{check})$ is the only part
of $\Pi_{solve} \setminus \Pi_{guess}$ where literals from
$Lit(\Pi_{guess})$ occur, we obtain
\begin{eqnarray*}
\Pi'_S &=&  F(\Pi_{check}[S]) \cup A_S \cup \Pi_{meta} \cup \{\derives\
\dneg{\texttt{notok}}.\}\\
       &=& \tr{\Pi_{check}[S]} \cup A_S \cup \{\derives\
\dneg{\texttt{notok}}.\}.
\end{eqnarray*}
The additional facts $A_S$ can be viewed as independent part of any
answer set of $\Pi'_S$, since the answer sets of
$\Pi'_S$ are the sets $T \cup A_S$ where $T$ is any answer set of
$\Pi'_S \setminus A_S$; note that $T\cap A_S = \emptyset$.
Indeed, the only rule in $\Pi'_S$ where the facts of $A_S$ play a role,
is line 17 of $\Pi_{meta}$.
All ground instances of line 17 are of the following form:
\[
\texttt{\ \ \ notok \derives inS(l), inS(nl), l\kuneq nl,
atom(l,|l|), atom(nl,|l|).}
\]
We assume $r$ fires and $\texttt{atom(l,|l|)} \in A_S$
(resp.\ $\texttt{atom(nl,|l|} \in A_S$).
Then, in order for the rule to fire, \texttt{inS(l)}
(resp.\ \texttt{inS(nl)}) has to be true. However, this can only be the case
for literals \texttt{l} (resp.\ \texttt{nl}) occurring in a rule head
of $\Pi_{check}[S]$ (backwards, by the rules in line
15, 14 and 1 of $\Pi_{meta}$ and by definition of $\Pi'_{check}$), which
contradicts our assumption that $\texttt{atom(l,|l|)} \in A_S$
(resp.\ $\texttt{atom(nl,|l|} \in A_S$).
 Therefore, the facts of $A_S$ do not affect the rule in line 17
and consequently $\Pi'_S$ has an answer set if and only if
$\Pi'_S \setminus A_S$ has an answer set and these answer sets coincide
on $Lit(\Pi'_S) \setminus A_S$.

By Theorem~\ref{theo:corr}, we know that (i) $\tr{\Pi_{check}[S]}$
always has an answer set and (ii)  $\tr{\Pi_{check}[S]}$ has any answer set containing
\texttt{notok} (which is unique) if and only if  $\Pi_{check}[S]$ has
no answer set. However, the constraint
$\derives~\dneg{\texttt{notok}}.$ only allows for answer sets of
$\Pi'_S$ containing \texttt{notok}. Hence, an answer set
$S_{check}$
of $\Pi'_S \setminus A_S$ exists if and only if
$\Pi_{check}[S]$ has no answer set, equivalently, $\Pi_{check}\cup S$
has no answer set.

Conversely, suppose $S$ is an answer set of $\Pi_{guess}$ such that
$\Pi_{check}\cup S$ has no answer set; equivalently, $\Pi_{check}[S]$
has no answer set.  By Theorem~\ref{theo:corr}, we know that
$\tr{\Pi_{check}[S]} = F(\Pi_{check}[S]) \cup \Pi_{meta}$ has a unique
answer set $S_{check}$, and $S_{check}$ contains \texttt{notok}. Hence,
also the program
$$Q_S = F(\Pi_{check}[S]) \cup \Pi_{meta} \cup \{\derives\ \dneg{\texttt{notok}}.\}$$
 has the unique answer set
$S_{check}$.  On the other hand, since $S$ is an answer set of $\Pi_{guess}$ and
$Lit(\Pi_{guess})$ is a splitting set for $\Pi_{solve}$, for each
answer set $S''$ of the program $\Pi'_S = (\Pi_{solve} \setminus
\Pi_{guess})[S]$, we have that $S \cup S''$ is an answer set of
$\Pi_{solve}$. However, $\Pi'_S = Q_S \cup A_S$; hence,
$S''=S_{check} \cup A_S$ must
hold and $S_{solve} = S \cup S_{check} \cup A_S$  is the unique answer set of
$\Pi_{solve}$ which extends $S$. This proves the result.
\end{proof}

The optimizations \textbf{OPT$_{pa}$} and \textbf{OPT$_{dep}$}
in Section~\ref{sec:optimizations} still apply. However, concerning \textbf{OPT$_{mod}$}, the following modifications are necessary:
\begin{trivlist}
\item 1.\ Like the input representation,
rules (\ref{stmt:rulesatisfied}) and (\ref{stmt:forcenormal})
have to be extended by adding {\sf body}$_{guess}$\texttt{(r)}.

\item 2.\ As for constraints $c$, we mentioned above that the factual
representation
of literals in $\body{c}$ may be skipped. This now only applies to literals
in $\pbody{c}$; the rule \texttt{lit(n,$l,c$)\ \derives\ {\sf body}$_{guess}(c)$.}
for $l\in \nbody{c}$ may no longer be dropped in general, as shown
by the following example.

\begin{example}\rm
Let $\Pi_{guess} = \{\ \texttt{g \vel -g.}\ \}$ and
$\Pi_{check}=\{\ r1:\texttt{~x \derives g.},\ r2:\texttt{~\derives \dneg{x}.}\ \}$
The ``input''  representation of $\Pi_{check}$ with respect to optimization
\textbf{OPT$_{mod}$}, i.e., the variable part of $\Pi'_{check}$, now consists of:
{\tt
\begin{tabbing}
 \quad \ \ \ \ \=lit(h,x,$r1$)$\,$\derives{}$\,$g.\ \ lit(n,x,$r2$).\ \ inS(x)$\,$:-$\,$g.\ \ notok$\,$:-$\,$ninS(x).
\end{tabbing}
}
\noindent where the latter correspond to rules (\ref{stmt:forcenormal}) and
(\ref{stmt:rulesatisfied}).
If we now assume that we want to check answer set $S=\set{\,\mbox{\tt-g}
\,}$ of $\Pi_{guess}$,
it is easy to see that $\Pi_{check}$ has no answer set for $S$,
and therefore $S$ should be represented by some answer set of our integrated
encoding. Now assume that {\tt lit(n,x,$r2$).} is dropped and we proceed
in generating the integrated encoding as outlined above with respect to
\textbf{OPT$_{mod}$}.
Since $\texttt{g} \not\in S$ and we have dropped
{\tt lit(n,x,$r2$).}, the ``input'' representation
of $\Pi_{check}$ for $S$ comprises
only the final rule \texttt{notok :- ninS(x).}. However, this rule can never
fire because neither line~15 nor line~16 of $\Pi_{meta}$ can ever
derive {\tt ninS(c)}. Therefore, also \texttt{notok} can not be derived and
the integrated check fails. On the other hand, {\tt lit(n,x,$r2$).} suffices
to derive {\tt ninS(x)} from line~16 of $\Pi_{meta}$, such that \texttt{notok}
can be derived and the integrated check works as intended.
\end{example}

In certain cases, we can still drop $l\in \nbody{c}$. For example, if $l$
occurs in the head of a rule $r$ with {\sf body}$_{guess}(r) = \emptyset$,
since in this case {\tt lit(h,l,$r$)} will always be added to the
program (see also respective remarks in Section~\ref{sec:applications}).%
\end{trivlist}

\subsection{Integrating Guess and $\NP\!$ Check Programs}

In contrast to the situation above, integrating a guess program $\Pi_{guess}$ and a check
program $\Pi_{check}$ which succeeds iff $\Pi_{check}\cup S$ has \emph{some}
answer set, is easy. Given that $\Pi_{check}$ is a HDLP again, this amounts
to integrating a check which is in $\NP$. After a rewriting to ensure the Splitting Set
property (if needed), simply take $\Pi_{solve}=\Pi_{guess} \cup
\Pi_{check}$; its answer sets correspond on the predicates in
$\Pi_{guess}$ to the desired solutions.


\section{Applications}
\label{sec:applications}

We now give examples of the use of our transformation for three well-known
$\SigmaP{2}$-complete problems from the literature, which involve $\coNP$-complete
checking for a polynomial-time solution guess: the first is about quantified Boolean formulas (QBFs)
with one quantifier alternation, which are well-studied in Answer Set
Programming, the second about conformant planning
\cite{eite-etal-2001e,turn-2002,leon-etal-2001}, and the third
is about strategic companies in the business domain \cite{leon-etal-2002-dlv}.

Further examples and ad hoc encodings of such problems can be found e.g.\ in
\cite{eite-etal-97f,eite-etal-2002-tplp,leon-etal-2002-dlv}
(and solved similarly). However, note that our method is applicable
to {\em any} checks encoded by inconsistency of a HDLP;
$\coNP$-hardness is not a prerequisite.


\subsection{Quantified Boolean formulas}
\label{sec:QBF}

Given a QBF $F = \exists x_1 \cdots \exists x_m \forall y_1 \cdots
\forall y_n\,\Phi$, where $\Phi = c_1 \vee \cdots \vee c_k $ is a propositional formula over
$x_1,\ldots,x_m, y_1,\ldots,y_n$ in disjunctive normal form, i.e. each
$c_i = a_{i,1} \wedge \cdots \wedge a_{i,l_i}$ and $|a_{i,j}| \in
\{x_1,\ldots,x_m,y_1\ldots,y_n\}$, the problem is to compute some
resp.\ all assignments to the
variable $x_1,\ldots,x_m$ which witness that $F$ evaluates to true.

Intuitively, this problem can be solved by ``guessing and checking''
as follows:
\begin{description}
\item[($\mathit{QBF}_{guess}$)] Guess a truth assignment for the variables $x_1,\ldots, x_m$.
\item[($\mathit{QBF}_{check}$)] Check whether this (fixed) assignment satisfies $\Phi$ for all
assignments of variables $y_1, \ldots, y_n$.
\end{description}

\noindent Both parts can be encoded by very simple HDLPs (or similarly by normal programs):

{\tt\small
\begin{tabbing}
\mbox{}\hspace*{3cm}\=\kill
\> \mbox{$\mathit{QBF}_{guess}:$} \+\\[0.75ex]
   $x_1$ \vel\ $-x_1.$ \ldots\ $x_m$ \vel\ $-x_m.$ \\
[3ex]
 \mbox{$\mathit{QBF}_{check}:$}\\[0.75ex]
 $y_1$ \vel \= \ $-y_1.$ \ldots\ $y_n$ \vel \ $-y_n.$\\[0.75ex]
\> \derives $a_{1,1}, \ldots, a_{1,l_1}$.\\
\> \phantom{sas} \vdots\\
\> \derives $a_{k,1}, \ldots, a_{k,l_1}$.
\end{tabbing}}


Clearly, both programs are head-cycle free. Moreover, for every
answer set $S$ of $\mathit{QBF}_{guess}$ --representing an
assignment to $x_1,\ldots,x_m$-- the program
$\mathit{QBF}_{check} \cup S$ has no answer set thanks to
the constraints, iff every assignment for $y_1, \ldots, y_n$
satisfies formula $\Phi$.

\begin{figure}
\caption{Integrated encoding $QBF_{solve}$ for QBF
 $\exists x_0x_1\forall y_0y_1 (\tneg{x_0} \wedge \tneg{y_0}) \vee
(y_0 \wedge \tneg{x_0}) \vee (y_1 \wedge x_0 \wedge \tneg{y_0}) \vee (y_0 \wedge \tneg{x_1} \wedge \tneg{y_0})$
}\label{fig:qbf}

{\footnotesize
\begin{alltt}
%%%% GUESS PART
 x0 v -x0.  x1 v -x1.

%%%% REWRITTEN CHECK PART
%% 1. Create dynamically the facts for the check program:

% y0 v -y0.                               % y1 v -y1.
  lit(h,"y0",1). lit(h,"-y0",1).            lit(h,"y1",2). lit(h,"-y1",2).
  atom("y0","y0"). atom("-y0","y0").        atom("y1","y1"). atom("-y1","y1").

 % :- -y0, -x0.
 % :-  y0, -x0.
 % :- -y0, y1, x0.
 % :- -y0, y0, -x1.

%% 2. Optimized meta-interpreter
%% 2.1 -- program dependent part

  notok :- ninS("y0"),ninS("-y0").
  notok :- ninS("y1"),ninS("-y1").
  notok :- inS("-y0"),-x0.
  notok :- inS("y0"),-x0.
  notok :- inS("y1"),inS("-y0"),x0.
  notok :- inS("y0"),inS("-y0"),-x1.

%% 2.2 -- fixed rules

% Iterate only over rules which contain L in the head:
  rule(L,R) :- lit(h,L,R), not lit(p,L,R), not lit(n,L,R).
  ruleBefore(L,R) :- rule(L,R), rule(L,R1), R1<R.
  ruleAfter(L,R) :- rule(L,R), rule(L,R1), R<R1.
  ruleBetween(L,R1,R2) :- rule(L,R1), rule(L,R2), rule(L,R3), R1<R3, R3<R2.
  firstRule(L,R) :- rule(L,R), not ruleBefore(L,R).
  lastRule(L,R) :- rule(L,R), not ruleAfter(L,R).
  nextRule(L,R1,R2) :- rule(L,R1), rule(L,R2), R1<R2, not ruleBetween(L,R1,R2).

% hlits are only those from active rules:
  hlit(L) :- rule(L,R).
  inS(L) v ninS(L) :- hlit(L).
  ninS(L) :- lit(HPN,L,R), not hlit(L).

% Consistency check could be skipped for programs without class. negation:
  notok :- inS(L), inS(NL), L != NL, atom(L,A), atom(NL,A).

  dep(L,L1) :- rule(L,R),lit(p,L1,R),inS(L1), inS(L).
  dep(L,L2) :- rule(L,R),lit(p,L1,R),dep(L1,L2),inS(L).
  cyclic :- dep(L,L1), dep(L1,L).
  phi(L,L1) v phi(L1,L) :- dep(L,L1), dep(L1,L), L<L1, cyclic.
  phi(L,L2) :- phi(L,L1),phi(L1,L2), cyclic.
  failsToProve(L,R) :- rule(L,R), lit(p,L1,R), ninS(L1).
  failsToProve(L,R) :- rule(L,R), lit(n,L1,R), inS(L1).
  failsToProve(L,R) :- rule(L,R), rule(L1,R), inS(L1), L1\kuneq{}L.
  failsToProve(L,R) :- lit(p,L1,R), rule(L,R), phi(L1,L), cyclic.
  allFailUpto(L,R) :- failsToProve(L,R), firstRule(L,R).
  allFailUpto(L,R1) :- failsToProve(L,R1), allFailUpto(L,R), nextRule(L,R,R1).
  notok :- allFailUpto(L,R), lastRule(L,R), inS(L).
  phi(L,L1) :- notok, hlit(L), hlit(L1), cyclic.
  inS(L) :- notok, hlit(L).
  ninS(L) :- notok, hlit(L).

%%% 3. constraint
   :- not notok.
\end{alltt}}
\end{figure}

By the method described in Section~\ref{sec:integration}, we can automatically generate a
single program $\Pi_{solve}$ integrating the guess and check programs.
For illustration, we consider the following QBF:
$$
\exists x_0x_1\forall y_0y_1
(\tneg{x_0} \wedge \tneg{y_0}) \vee
(y_0 \wedge \tneg{x_0}) \vee
(y_1 \wedge x_0 \wedge \tneg{y_0}) \vee
(y_0 \wedge \tneg{x_1} \wedge \tneg{y_0})
$$
This QBF evaluates to true: for the assignments $x_0=0,x_1=0$  and
$x_0=0,x_1=1$, the subformula $\forall y_0y_1(\cdots)$ is a
tautology.

The integrated program $\mathit{QBF}_{solve} = \mathit{QBF}_{guess} \cup \mathit{QBF}'_{check}$
under use of the optimized transformation $\trOpt{\cdot}$ of $\trP(\cdot)$ as
discussed is shown in Figure~\ref{fig:qbf}. It has two answer
sets of the form $S_1 = \{ x_0,  -x_1, \ldots, \}$ and
$S_2 = \{ x_0,  x_1, \ldots, \}$, respectively.

With respect to the variants of the transformation, we remark that for
the QBF encoding considerations upon negative literals in constraints
in \textbf{OPT$_{mod}$} do not play a role, because all literals in
the constraints of $\mathit{QBF}_{check}$ are positive. Also
\textbf{OPT$_{pa}$} does not play a role, since the only rules in
$\mathit{QBF}_{check}$ with non-empty heads are always potentially
applicable because their bodies are empty.

Note that the customary (but tricky) saturation technique in
disjunctive logic programming to solve this problem, as used
e.g.\ in \cite{eite-etal-97f,leon-etal-2002-dlv} and
shown in 
\ref{app:adhoc-qbf},  is fully transparent to the
non-expert, who might easily come up with the two programs above.

\subsection{Conformant planning}
\label{sec:conformant}

Loosely speaking, planning is the problem of finding a sequence of
actions $P=\alpha_1$, $\alpha_2$,\ldots, $\alpha_n$, a {\em plan},
which takes a system from an initial state $s_0$ to a state $s_n$ in
which a goal (often, given by an atom $g$) holds, where a state $s$ is
described by values of fluents, i.e., predicates which might change
over time. {\em Conformant planning} \cite{gold-bodd-96} is concerned with
finding a plan $P$ which works under all contingencies which may
arise because of incomplete information about the initial state and/or
nondeterministic action effects.

As well-known, conformant planning in a STRIPS-style formulation is a
$\SigmaP{2}$-complete problem (precisely, deciding plan existence) in
certain settings, e.g.\ if the plan length $n$ (of polynomial size) is
given and executability of actions is guaranteed, cf.\
\cite{eite-etal-2001e,turn-2002}. Hence, the problem can be solved
with a guess and ($\coNP$) check strategy.

As an example, we consider
a simplified version of the well-known ``{\em Bomb in the Toilet}''
planning problem \cite{mcde-87} as in \cite{eite-etal-2001e}: We have
been alarmed that a possibly armed bomb is in a lavatory which has a
toilet bowl. Possible actions are dunking the bomb into the bowl and
flushing the toilet. After just dunking, the bomb may be disarmed or
not; only flushing the toilet guarantees that it is really disarmed.

Using the following guess and
check programs $\mathit{Bomb_{guess}}$ and $\mathit{Bomb_{check}}$,
respectively, we can compute a plan for having the bomb disarmed by
two actions:%

\vspace{.125ex}

{\tt\small
\begin{tabbing}
X\qquad\quad\=\kill \>\mbox{$Bomb_{guess}:$} \+\\[0.75ex]
\% Timestamps: \\
time(0). time(1). \\[0.5ex]
\% Guess a plan:\\
dunk(T) v -dunk(T) :- time(T).   \\
flush(T) v -flush(T) :- time(T). \\[0.5ex]
\% Forbid concurrent actions:    \\
:- flush(T), dunk(T).            \\[3ex]
\mbox{$Bomb_{check}:$}\\[0.75ex]
\% Initial state: \\
 armed(0) v -armed(0).\\[0.5ex]
\% Frame Axioms:\\
armed(T1) :- armed(T), not -armed(T1), time(T), T1=T+1.\\
dunked(T1) :- dunked(T), T1=T+1.\\[0.5ex]
\% Effect of dunking:\\
dunked(T1) :- dunk(T), T1=T+1.\\
armed(T1) v -armed(T1) \derives dunk(T), armed(T), T1=T+1.\\[0.5ex]
\% Effect of flushing:\\
-armed(T1) :- flush(T), dunked(T), T1=T+1.\\[0.5ex]
\% Check whether goal holds in stage 2: \\
 :- not armed(2). \-
\end{tabbing}}

\vspace{.125ex}

\noindent
$\mathit{Bomb_{guess}}$ guesses all candidate plans
$P=\alpha_1,\alpha_2$, starting from possible time points for action execution,
while  $\mathit{Bomb_{check}}$ checks whether
any such plan $P$ is conformant for the goal $g$ =
\texttt{not armed(2)}. Here, the closed world assumption (CWA) on
\texttt{armed} is used, i.e., absence of \texttt{armed($t$)} is viewed as
\texttt{-armed($t$)}, which saves a negative frame axiom on
\texttt{-armed}. The final constraint eliminates a plan execution
iff it reaches the goal; thus, $\mathit{Bomb_{check}}$ has
no answer set iff the plan $P$ is conformant.  As can be checked,
the answer set $S=\{ \texttt{time(0)}, \texttt{time(1)},
\texttt{dunk(0)}, \texttt{flush(1)}\}$ of $\mathit{Bomb_{guess}}$
corresponds to the (single) conformant plan $P$= \texttt{dunk},
\texttt{flush} for the goal \texttt{not armed(2)}.

By using the method from Section~\ref{sec:integration}, the programs
$\mathit{Bomb_{guess}}$ and  $\mathit{Bomb_{check}}$ can be integrated
automatically into a single program $\mathit{Bomb_{plan}}= \mathit{Bomb_{guess}} \cup
\mathit{Bomb_{check}}'$
(cf.\ 
\ref{app:planning}). It has a single answer set, corresponding to
the single conformant plan $P$ = \texttt{dunk}, \texttt{flush} as desired.

We point out that our rewriting method is more generally applicable
than the encoding for conformant planning proposed in \cite{leon-etal-2001}. It loosens some of the restrictions
there: While \cite{leon-etal-2001} requires that the state transition
function is specified by a positive constraint-free logic program, our
method can still safely be used in presence of negation and
constraints, provided action execution will always lead to a
consistent successor state and not entail absurdity; see
\cite{eite-etal-2001e,turn-2002} for a discussion of this setting.

Concerning \textbf{OPT$_{mod}$}, we point out that there is the
interesting constraint
\[
\texttt{$c:$ :- not armed(2).}
\]
in program $\mathit{Bomb_{check}}$. Here, we may drop
$\texttt{lit(h,"armed(2)",c)}$ safely: For the frame axiom
\[
\texttt{$r:$ armed(2)\ \derives\ armed(1),\ not\ -armed(2),\ time(1).}
\]
(cf.\ 
\ref{app:planning}), we have {\sf body}$_{guess}(r) =
\set{\texttt{time(1)}}$. Therefore, we obtain:
\[
\texttt{lit(h,"armed(2)",r) :- time(1).}
\]
However, this rule will always be added since $\texttt{time(1)}$ is a
deterministic consequence of $\mathit{Bomb_{guess}}$.  As for
\textbf{OPT$_{pa}$} and considering the ``Bomb in the Toilet''
instances from \cite{eite-etal-2001e}, there might be rules which are
\emph{not} possible applicable with respect to a guessed plan;
however, in experiments, the additional overhead for computing
unfounded sets did not pay off.

A generalization of the method demonstrated here on a small planning
problem expressed in Answer Set Programming to conformant planning in
the \dlvk{} planning system \cite{eite-etal-2001e}, is discussed in
detail in~\cite{poll-2003}.  In this system, planning problems are
encoded in a logical action language, and the encodings are mapped to
logic programs. For conformant planning problems, separate guess and
check programs have been devised \cite{eite-etal-2001e}, which by our
method can be automatically integrated into a single logic
program. Such an encoding was previously unkown.

\subsection{Strategic Companies}
\label{sec:strategic}

Another $\SigmaP{2}$-complete problem is the strategic companies
problem from \cite{cado-etal-97}. Briefly, a holding owns companies,
each of which produces some goods.  Moreover, several companies may
jointly have control over another company. Now, some companies should
be sold, under the constraint that all goods can be still produced,
and that no company is sold which would still be controlled by the
holding after the transaction. A company is {\em strategic}, if it
belongs to a {\em strategic set}, which is a minimal set of companies
satisfying these constraints.  Guessing a strategic set, and checking
its minimality can be done by the following two programs, where we
adopt the constraint in \cite{cado-etal-97} that each product is
produced by at most two companies and each company is jointly
controlled by at most three other companies.

{\tt\small
\begin{tabbing}
XX\= \kill \>\mbox{$SC_{guess}:$} \+\\[.75ex]
strat(X) \= v -strat(X)  \derives company(X). \\[.5ex]
\>                      \derives prod\_by(X,Y,Z), not strat(Y), not strat(Z). \\[.5ex]
\>                      \derives \= contr\_by(W,X,Y,Z), not strat(W),\\
                               \>\> strat(X), strat(Y), strat(Z).\\[2ex]
\mbox{$SC_{check}:$}\\[.75ex]
strat1(X) \= v -strat1(X) \derives strat(X).\\[.5ex]
\> \derives prod\_by(X,Y,Z), not strat1(Y), not strat1(Z).\\[.5ex]
\> \derives \= contr\_by(W,X,Y,Z), not strat1(W),\\
\>\>  strat1(X), strat1(Y), strat1(Z).\\[.5ex]
\> smaller \' \derives\ -strat1(X).\\[.5ex]
\> \derives not smaller.
\end{tabbing}}

Here, {\tt strat$(C)$} means that $C$ is strategic, {\tt
prod\_by$(P,C1,C2)$} that product $P$ is produced by
companies $C1$ and $C2$, and {\tt contr\_by}$(C,C1,$
$C2,C3)$ that $C$ is jointly controlled by
$C1,C2$ and $C3$.
We assume facts \texttt{company($\cdot$)., prod\_by($\cdot,\cdot,\cdot$).}, and
\texttt{contr\_by($\cdot,\cdot,\cdot,\cdot$).} to be defined in a
separate program which can be considered as part of $SC_{guess}$.

 The two programs above intuitively encode guessing a set \texttt{strat} of
companies which fulfills the production and control preserving constraints,
such that no real subset \texttt{strat1} fulfills these constraints.
While the ad hoc encodings from \cite{eite-etal-2000c,leon-etal-2002-dlv},
which can also be found in 
\ref{app:adhoc-sc}, are not immediate
(and require some thought), the above programs are very natural and easy
to come up with.

\begin{table}
\begin{center}
\small
\begin{tabular}[t]{|c|c|c|}
\cline{1-3}
PRODUCT & COMPANY \#1 & COMPANY \#2 \\
\cline{1-3}
 Pasta & Barilla & Saiwa\\
 Tomatoes & Frutto & Barilla\\
 Wine & Barilla & -- \\
 Bread & Saiwa & Panino \\
\cline{1-3}
\end{tabular}
\caption{Relation $\mathit{prod\_by}$ storing producers of each good}\label{tab:prod}
\end{center}
\end{table}

As an example, let us consider the following production and control
relations from \cite{cado-etal-97} in a holding as shown in Tables \ref{tab:prod} and
\ref{tab:contr}. The symbol ``--'' there means that the entry is void,
which we simply represent by duplicating the single producer (or one of the controlling companies, respectively)
in the factual representation; a possible representation is thus

{\alltt\small
    company(barilla). company(saiwa).
    company(frutto).  company(panino).
    prod\_by(pasta,barilla,saiwa).  prod\_by(tomatoes,frutto,barilla).
    prod\_by(wine,barilla,barilla). prod\_by(bread,saiwa,panino).
    contr\_by(frutto,barilla,saiwa,saiwa).
}

\medskip

If we would consider only the production relation, then Barilla and
Saiwa together would form a strategic set,
because they jointly produce all goods
but neither of them alone. On the other hand, Frutto would not be strategic.
However, given the company control as in Table~\ref{tab:contr} means that
Barilla and Saiwa together have control over Frutto.
Taking into account that therefore Frutto can be sold only if either Barilla or Saiwa
is also sold, the minimal sets of companies that produce all goods
change completely: $\{$Barilla, Saiwa$\}$
is no longer a strategic set, while $s_1=\{$Barilla, Saiwa, Frutto$\}$ is.
Alternatively, $s_2=\{$Barilla, Panino$\}$ is another strategic set.

\begin{table}
\begin{center}
\small
\begin{tabular}{|c|c|c|c|}
\cline{1-4}
 CONTROLLED & CONT \#1 & CONT \#2 & CONT \#3 \\
\cline{1-4}
 Frutto & Barilla & Saiwa & --\\
\cline{1-4}
\end{tabular}
\caption{Relation $\mathit{contr\_by}$ storing company control information} \label{tab:contr}
\end{center}
\end{table}

Integration of the programs $SC_{guess}$ and $SC_{check}$ after
grounding is again possible by the method from
Section~\ref{sec:integration} in an automatic way. Here, the facts
representing the example instance are to be added as part of
$SC_{guess}$, yielding two answer sets corresponding to $s_1$ and
$s_2$ (cf.\ 
\ref{app:stratcomp}).

\medskip
With regard to \textbf{OPT$_{mod}$}, we remark that depending on the
concrete problem instance, $SC_{check}$ contains critical constraints
$c$, where \texttt{\dneg{strat1}($\cdot$)} occurs, such that
\texttt{lit(n,"strat1($\cdot$)",c)} may not be dropped
here  (cf.\ 
\ref{app:stratcomp}).
Furthermore, as for \textbf{OPT$_{pa}$} all rules with non-empty
heads are either possibly applicable or ``switched off'' by
$SC_{guess}$.  Since there are no positive dependencies among the
rules, \texttt{pa($\cdot$)} does not play a role there.

As a final remark, we note that modifying the guess and check programs
$SC_{guess}$ and $SC_{check}$ to allow for unbounded numbers of
producers for each product and controllers for each company,
respectively, is easy. Assume that production and control are
 represented instead of relations \texttt{prod\_by} and \texttt{contr\_by}
by an arbitrary number of facts of the form
\texttt{produces($c,p$).}  and
\texttt{controls($c_1,g,c$).}, which state that
company $c$ produces $p$ and that company $c_1$ belongs
to a group $g$ of companies which jointly control $c$,
respectively.  Then, we would simply have to change the constraints in
$SC_{guess}$ to:

{\tt
\begin{tabbing}
\ \ \ \ \=no\_control(G,C) :- controls(C1,G,C), not strat(C1).\\[0.33ex]
\>    :- controls(C1,G,C), not no\_control(G,C), not strat(C).\\[1ex]
\>    produced(P) :- produces(C,P), strat(C).\\[0.33ex]
\>    :- produces(C,P), not produced(P).
\end{tabbing}
}

\medskip

The constraints in $SC_{check}$ are changed similarly. Then, the
synthesized integrated encoding according to our method gives us a
DLP solving this problem. The ad hoc encodings  in 
\cite{eite-etal-2000c,leon-etal-2002-dlv} can not be adapted that easily,
and in fact require substantial changes. 

\section{Experiments}
\label{sec:experiments}


As for evaluation of the proposed approach we have conducted a series of
experiments for the problems outlined in the previous Section.
Here, we were mainly interested in the following questions:
\begin{myenumerate}
\item \emph{What is the performance impact of our automatically
generated, integrated encoding compared with ad hoc encodings of
$\SigmaP{2}$ problems?}

We have therefore compared our
automatically generated integrated encoding of QBFs and
Strategic Companies
against the following ad hoc encodings:

\begin{enumerate}
    \item[(i)] QBF against the ad hoc encoding for QBFs described
in \cite{leon-etal-2002-dlv} (which assumes that the quantifier-free
part is in 3DNF, i.e., contains three literals per disjunct); see
\ref{app:adhoc-qbf}.

  \item [(ii)] Strategic companies against the two ad hoc encodings
for the Strategic Companies problem from \cite{eite-etal-2000c};  see
\ref{app:adhoc-sc}.

These two encodings significantly differ: The first
encoding, $ad hoc_1$ is very concise, and integrates guessing and checking in only two
rules; it is an illustrative example of the power of disjunctive
rules and tailored for a DLP system under answer set semantics.
The second encoding, $ad hoc_2$, has a more obvious separate structure of the guessing
and checking parts of the problem at the cost of some extra rules.
However, in our opinion, none of these ad hoc encodings is obvious at
first sight compared with the separate guess and check programs shown above.

\end{enumerate}

 Concerning (i) we have tested randomly generated QBF instances with $n$
existentially and $n$ universally quantified variables (QBF-$n$), and
concerning (ii) we have chosen randomly generated instances involving
$n$ companies (SC-$n$).

\item \emph{What is the performance impact of the automatically
generated, integrated encoding compared with interleaved computation
of guess and check programs?}

To this end, we have tested the
performance of solving some conformant planning problems
with integrated encodings compared with the ASP based planning
system \dlvk\ \cite{eite-etal-2001e} which solves conformant planning
problems by interleaving the guess of a plan with checking plan security.
For its interleaved computation, \dlvk hinges on translations of the
planning problem to HDLPs, by computing ``optimistic'' plans as solutions
of a HDLP $\Pi^{plan}_{guess}$ and interleaved checking of plan security by
non-existence of solutions of a new program $\Pi^{plan}_{guess}$
which is dynamically generated with respect to
the plan at hand. \dlvk generalizes in some sense solving the
small planning example in Section~\ref{sec:conformant} for arbitrary
planning problems specified in a declarative language,
\K \cite{eite-etal-2001d}. For our experiments we have used elaborations
of ``Bomb in the Toilet'' as described in \cite{eite-etal-2001d}, namely
``Bomb in the Toilet with clogging'' BTC($i$), where the toilet is clogged
after dunking a package, and ``Bomb in the Toilet with Uncertain Clogging''
BTUC($i$) where this clogging effect is non-deterministic and there
are $i$ many possibly armed packages.
\end{myenumerate}

\subsection{Test Environment and General Setting}
All tests were performed on an AMD Athlon 1200MHz machine with 256MB of
main memory running SuSE Linux 8.1.

All our experiments have been conducted using the
\dlv{} system \cite{leon-etal-2002-dlv,dlv-web}, which is a state-of-the-art
Answer Set Programming engine capable of solving DLPs. Another available
system, \gnt~\cite{janh-etal-2000}\footnote{\gnt, available from \url{http://www.tcs.hut.fi/Software/gnt/}, is an extension of \smodels
solving DLPs by interleaved calls of \smodels, which itself is only
capable of solving normal LPs.} which is not reported here showed
worse performance/higher memory consumption on the tested instances.

 Since our method works on ground programs, we had to ground all instances
(i.e.\ the corresponding guess and check programs) beforehand whenever
dealing with non-ground programs. Here, we have used \dlv grounding
with most optimizations turned off:\footnote{Respective ground instances
have been produced with the command \texttt{dlv -OR- -instantiate},
(cf.\ the \dlv-Manual \cite{dlv-web}), which turns off most
of the grounding optimizations.} Some optimizations during \dlv grounding
rewrite the program, adding new predicate symbols, etc. which we
turned off in order to obtain correct input for the meta-interpreters.

In order to assess the effect of various optimizations and
improvements to the transformation \tr{\cdot}, we have also conducted the
above experiments with the integrated encodings based on different
optimized versions of \tr{\cdot}.

\subsection{Results}

The results of our experiments are shown in Tables~\ref{tab:QBF}-\ref{tab:BIT}.
We report there the following tests on the various instances:
\begin{itemize}
\item $meta$\  indicates the unoptimized meta-interpreter $\Pi_{meta}$
\item $mod$\ \ indicates the non-modular optimization \textbf{OPT$_{mod}$} including the refinement
for constraints.
\item $dep$\quad indicates the optimization \textbf{OPT$_{dep}$} where \inlinek{phi} is only guessed for literals mutually depending on each other through
positive recursion.
\item $opt$\quad indicates both optimizations \textbf{OPT$_{mod}$} and
\textbf{OPT$_{dep}$} turned on.
\end{itemize}

We did not include optimization \textbf{OPT$_{pa}$} in our
experiments, since the additional overhead for computing unfounded rules
in the check programs which we have considered did not pay off (in fact,
\textbf{OPT$_{pa}$} is irrelevant for QBF and Strategic
Companies).%

All times reported in the tables represent the execution times for
finding the first answer set under the following resource constraints.
We set a time limit of 10 minutes
(=600 seconds) for QBFs and Strategic Companies, and of 4.000 seconds for
the ``Bomb in the Toilet'' instances. Furthermore, the limit on memory
consumption was 256 MB (in order to avoid swapping).  A dash '-' in the tables
indicates that one or more instances exceeded these limits.

\renewcommand{\arraystretch}{1.2}

\begin{sidewaystable}

\begin{centering}


{\footnotesize
\renewcommand{\arraystretch}{1.125}
\begin{tabular}{|@{\ }l@{\ }|*{10}{c|}}
\cline{1-11}
\multicolumn{1}{|c|}{}&         \multicolumn{2}{c|}{$ad hoc$~\cite{leon-etal-2002-dlv}} &
          \multicolumn{2}{c|}{$meta$} &
          \multicolumn{2}{c|}{$mod$} &
          \multicolumn{2}{c|}{$dep$} &
          \multicolumn{2}{c|}{$opt$}  \\

\multicolumn{1}{|c|}{}& AVG   & MAX   & AVG   & MAX   & AVG   & MAX   & AVG   & MAX   & AVG   & MAX \\
\cline{1-11}
 QBF-4    & 0.01s & 0.02s & 0.16s & 0.18s & 0.10s & 0.15s & 0.09s & 0.11s & 0.07s & 0.09s\\
 QBF-6    & 0.01s & 0.02s & 1.11s & 1.40s & 0.25s & 1.12s & 0.17s & 0.21s & 0.08s & 0.12s\\
 QBF-8    & 0.01s & 0.06s & 10.4s & 16.3s & 1.18s & 7.99s & 0.49s & 0.87s & 0.10s & 0.23s\\
 QBF-10   & 0.02s & 0.09s & 82.7s & 165s  & 4.34s & 30.7s & 1.74s & 3.67s & 0.12s & 0.36s\\
 QBF-12   & 0.02s & 0.16s &   - &  -  &   - &  -  &   - &  -  & 0.15s & 0.79s\\
 QBF-14   & 0.06s & 1.21s &   - &  -  &   - &  -  &   - &  -  & 0.34s & 5.87s\\
 QBF-16   & 0.08s & 1.85s &   - &  -  &   - &  -  &   - &  -  & 0.44s & 10.3s\\
 QBF-18   & 0.19s & 7.12s &   - &  -  &   - &  -  &   - &  -  & 1.04s & 38.8s\\
 QBF-20   & 1.49s & 21.3s &   - &  -  &   - &  -  &   - &  -  & 7.14s & 101s\\
\cline{1-11}
\multicolumn{11}{c}{}\\[-1ex]
\multicolumn{11}{c}{\small\rm Average and maximum times for 50 randomly chosen instances per size.}
\end{tabular}
}

\vspace*{-1ex}

\caption{Experiments for  QBF} \label{tab:QBF}

\bigskip
\bigskip

{\footnotesize
\renewcommand{\arraystretch}{1.125}

\begin{tabular}{|l|c|c|c|c|c|c|c|c|c|c|c|c|}
\cline{1-13}
\multicolumn{1}{|c|}{}&
\multicolumn{2}{c|}{$ad hoc_1$~\cite{eite-etal-2000c}} &
\multicolumn{2}{c|}{$ad hoc_2$~\cite{eite-etal-2000c}} &
          \multicolumn{2}{c|}{$meta$} &
          \multicolumn{2}{c|}{$mod$} &
          \multicolumn{2}{c|}{$dep$} &
          \multicolumn{2}{c|}{$opt$}  \\
\multicolumn{1}{|c|}{}
              & AVG   & MAX   & AVG   & MAX   & AVG   & MAX   & AVG   & MAX   & AVG   & MAX   & AVG   & MAX \\
\cline{1-13}
SC-10  & 0.01s & 0.02s & 0.05s & 0.05s & 0.66s & 0.69s & 0.49s & 0.51s & 0.36s & 0.38s & 0.13s & 0.15s\\
SC-15  & 0.01s & 0.02s & 0.11s & 0.13s & 1.82s & 3.23s & 1.50s & 3.12s & 0.64s & 0.68s & 0.20s & 0.22s\\
SC-20  & 0.02s & 0.02s & 0.26  & 0.27s & 3.75s & 3.90s & 3.34s & 3.61s & 1.07s & 1.13s & 0.26s & 0.27s\\
SC-25  & 0.02s & 0.02s & 0.51s & 0.54s &   -   &   -   &   -   &   -   & 1.63s & 1.68s & 0.33s & 0.35s\\
SC-30  & 0.02s & 0.03s & 0.91s & 0.97s &   -   &   -   &   -   &   -   & 2.35s & 2.47s & 0.42s & 0.44s\\
SC-35  & 0.02s & 0.03s & 1.50s & 1.60s &   -   &   -   &   -   &   -   & 3.17s & 3.27s & 0.54s & 0.56s\\
SC-40  & 0.03s & 0.03s & 2.52s & 2.70s &   -   &   -   &   -   &   -   & 4.25s & 4.43s & 0.68s & 0.71s\\
SC-45  & 0.03s & 0.04s & 4.503 & 4.97s &   -   &   -   &   -   &   -   & 5.46s & 5.77s & 0.84s & 0.90s\\
SC-50  & 0.03s & 0.04s & 8.38s & 8.68s &   -   &   -   &   -   &   -   & 6.73s & 6.86s & 1.00s & 1.02s\\
SC-60  & 0.04s & 0.05s & 22.6s & 24.3s &   -   &   -   &   -   &   -   & 10.2s & 10.6s & 1.47s & 1.53s\\
SC-70  & 0.04s & 0.05s & 44.2s & 48.1s &   -   &   -   &   -   &   -   & 14.7s & 15.4s & 2.05s & 2.10s\\
SC-80  & 0.04s & 0.05s & 75.9s & 82.5s &   -   &   -   &   -   &   -   & 19.7s & 21.0s & 2.78s & 3.05s\\
SC-90  & 0.05s & 0.06s &  125s &  130s &   -   &   -   &   -   &   -   & 26.8s & 27.6s & 3.67s & 3.85s\\
SC-100 & 0.06s & 0.08s &  196s &  208s &   -   &   -   &   -   &   -   & 34.8s & 36.3s & 4.70s & 4.80s\\
\cline{1-13}
\multicolumn{13}{c}{}\\[-1ex]
\multicolumn{13}{c}{\rm\small Average and maximum times for 10 randomly chosen instances per size.}
\end{tabular}
}

\vspace*{-1ex}

\caption{Experiments for Strategic Companies} \label{tab:SC}
\end{centering}

\end{sidewaystable}

\begin{table}
\begin{centering}
{\footnotesize
\renewcommand{\arraystretch}{1.2}
\begin{tabular}{|@{\ }l@{\ }|c|c|c|c|c|}
\cline{1-6}
\multicolumn{1}{|c|}{}& {{\texttt{DLV}$^\K$}}\cite{eite-etal-2001e}
                & $meta$ & $mod$ &  $dep$ & $opt$\\
\cline{1-6}
BTC(2)  & 0.01s &  1.16s & 0.80s & 0.15s & 0.08s\\
BTC(3)  & 0.11s &  9.33s & 9.25s & 8.18s & 4.95s\\
BTC(4)  & 4.68s &  71.3s & 67.8s &  333s &  256s\\
\cline{1-6}
BTUC(2) & 0.01s &  6.38s & 6.26s & 0.22s & 0.17s\\
BTUC(3) & 1.78s &    - &  -  & 28.1s & 13.0s\\
BTUC(4) &  577s &    - &  -  &  -  & 2322s\\
\cline{1-6}
\multicolumn{6}{l}{}\\
\multicolumn{6}{c}{\small BTC, BTUC with 2,3 and 4 packages.}
\end{tabular}
}

\caption{Experiments for Bomb in Toilet}\label{tab:BIT}
\end{centering}
\end{table}

The results in Tables~\ref{tab:SC}-\ref{tab:BIT} show that the ``guess
and saturate'' strategy in our approach benefits a lot from
optimizations for all problems considered.  However, we emphasize that
it might depend on the structure of $\Pi_{guess}$ and $\Pi_{check}$
which optimizations are beneficial.  We
strongly believe that there is room for further improvements both on
the translation and for the underlying \dlv engine.

We note the following observations: 

\begin{itemize}
    \item Interestingly, for the QBF problem, the performance of our optimized
translation stays within reach of the ad hoc encoding in
\cite{leon-etal-2002-dlv} for small instances. Overall, the
performance shown in Table~\ref{tab:QBF} is within roughly a factor of
5-6 (with few exceptions for small instances), and thus scales similarly.

\item For the Strategic Companies problem, the picture in Table~\ref{tab:SC} is
 even more interesting. Unsurprisingly, the automatically generated encoding is inferior
to the succinct ad hoc encoding $ad hoc_1$; it is more than an
order of magnitude slower and scales worse. However, while it is
slower by a small factor than the ad hoc encoding $ad hoc_2$ (which is
 more involved) on small instances, it scales much better and quickly
outperforms this encoding.

\item For the planning problems, the integrated encodings tested still stay behind the interleaved
calls of $\dlvk$.

\item In all cases, the time limit was exceeded (for smaller
  instances) rather than the memory limit, but especially for bigger
  instances of ``Bomb in the toilet'' and ``Strategic Companies,'' in
  some cases the memory limit was exceeded before timeout (e.g.\ for BTUC(5), even with the optimized version of
  our transformation).

\end{itemize}

\section{Summary and Conclusion}

\label{sec:conclusion}

We have considered the problem of integrating separate ``guess'' and
``check'' programs for solving expressive problems in the Answer Set
Programming paradigm with a 2-step approach, into a single logic
program. To this end, we have first presented a polynomial-time
transformation of a head-cycle free, disjunctive program $\Pi$ into a
disjunctive program $\tr{\Pi}$ which is stratified and
constraint-free, such that in the case where $\Pi$ is inconsistent
(i.e., has no answer set), $\tr{\Pi}$ has a single designated answer
set which is easy to recognize, and otherwise the answer sets of $\Pi$
are encoded in the answer sets of $\tr{\Pi}$. We then showed how to
exploit $\tr{\Pi}$ for combining a ``guess'' program $\Pi_{solve}$ and
a ``check'' program $\Pi_{check}$ for solving a problem in Answer Set
Programming automatically into a single disjunctive logic program,
such that its answer sets encode the solutions of the problem.

Experiments have shown that such a synthesized encoding has weaker
performance than the two-step method or an optimal ad hoc encoding
for a problem, but can also outperform (reasonably looking) ad hoc
encodings. This is noticeable since in some cases, finding any
arbitrary ``natural'' (not necessarily optimal) encoding of a problem
in a single logic program appears to be very difficult, such as e.g.,\
for conformant planning \cite{leon-etal-2001} or determining minimal
update answer sets \cite{eite-etal-2002-tplp}, where such encodings
were not known for the general case.

Several issues remain for being tackled in future work. The first
issue concerns extending the scope of programs which can be handled.
The rewriting method which we have presented here applies to
propositional programs only.  Thus, before transformation, the program
should be instantiated.  In \cite{leon-etal-2002-dlv} instantiations
of a logic program used in \dlv have been described, which keep the
grounding small and do not necessarily ground over the whole Herbrand
universe.  For wider applicability and better scalability of the
approach, a more efficient lifting of our method to non-ground
programs is needed. Furthermore, improvements to the current
transformations might be researched. Some preliminary experimental
results suggest that a structural analysis of the given guess and
check program might be valuable for this purpose.

A further issue are alternative transformations, which are possibly
tailored for certain classes of programs.  The work of Ben-Eliyahu and
Dechter~\citeyear{bene-dech-94}, on which we build, aimed at
transforming head-cycle free disjunctive logic programs into SAT
problems. It might be interesting to investigate whether related
methods such as the one developed for ASSAT~\cite{lin-zhao-2002},
which was recently generalized by Lee and
Lifschitz~\citeyear{lee-lifs-2003} to disjunctive programs, can be
adapted for our approach.


\subsection*{Acknowledgments}

We thank Gerald Pfeifer for his help on experimental evaluation and
fruitful discussions. We are also grateful to the reviewers of the
paper as well as the reviewers of the preliminary conference versions
for their comments and constructive suggestions for improvement.

\appendix

\section{
Integrated Program for Conformant Planning}


\label{app:planning}

The integrated program for the planning problem
in Section~\ref{sec:conformant}, $\mathit{Bomb_{plan}} = \mathit{Bomb_{guess}} \cup
\mathit{Bomb_{check}}'$,  is given below.
It has a single answer set $S = \{$ \texttt{dunk(0)}, \texttt{-flush(0)},
\texttt{flush(1)}, \texttt{-dunk(1)}, \ldots $\}$ which corresponds to the single conformant plan $P$= \texttt{dunk},
\texttt{flush} as desired.

\medskip

{\footnotesize
\begin{alltt}
%%%% GUESS PART

% Timestamps:
  time(0). time(1).

% Guess a plan:
  dunk(T) v -dunk(T) :- time(T).
  flush(T) v -flush(T) :- time(T).
  :- flush(T), dunk(T).

%%%% REWRITTEN CHECK PART (after grounding)
%% 1. Create dynamically the facts for the program:

% armed(0) v -armed(0).
   lit(h,"armed(0)",1). atom("armed(0)","armed(0)").
   lit(h,"-armed(0)",1). atom("-armed(0)","armed(0)").

% armed(T1) :- armed(T), not -armed(T1), time(T), T1=T+1.
   lit(h,"armed(1)",2) :- time(0). atom("armed(1)","armed(1)").
   lit(p,"armed(0)",2) :- time(0).
   lit(n,"-armed(1)",2) :- time(0).

   lit(h,"armed(2)",3) :- time(1).  atom("armed(2)","armed(2)").
   lit(p,"armed(1)",3) :- time(1).
   lit(n,"-armed(2)",3) :- time(1).

% dunked(T1) :- dunked(T), T1=T+1.
   lit(h,"dunked(1)",4).  atom("dunked(1)","dunked(1)").
   lit(p,"dunked(0)",4).

   lit(h,"dunked(2)",5). atom("dunked(2)","dunked(2)").
   lit(p,"dunked(1)",5).

% dunked(T1) :- dunk(T), T1=T+1.
   lit(h,"dunked(1)",6) :- dunk(0).

   lit(h,"dunked(2)",7) :- dunk(1).

% armed(T1) v -armed(T1) :- dunk(T), armed(T), T1=T+1.
   lit(h,"armed(1)",8) :- dunk(0).
   lit(h,"-armed(1)",8) :- dunk(0). atom("-armed(1)","armed(1)").
   lit(p,"armed(0)",8) :- dunk(0).

   lit(h,"armed(2)",9) :- dunk(1).
   lit(h,"-armed(2)",9) :- dunk(1). atom("-armed(2)","armed(2)").
   lit(p,"armed(1)",9) :- dunk(1).

% -armed(T1) :- flush(1), dunked(T),T1=T+1.
   lit(h,"-armed(1)",10) :- flush(0). lit(p,"dunked(0)",10) :- flush(0).

   lit(h,"-armed(2)",11) :- flush(1). lit(p,"dunked(1)",11) :- flush(1).

% :- not armed(2).

%% 2. Optimized meta-interpreter

%% 2.1 -- program dependent part

   notok :- ninS("armed(0)"), ninS("-armed(0)").
   inS("armed(1)") :- inS("armed(0)"), ninS("-armed(1)"), time(0).
   inS("armed(2)") :- inS("armed(1)"), ninS("-armed(2)"), time(1).
   inS("dunked(1)") :- inS("dunked(0)").
   inS("dunked(2)") :- inS("dunked(1)").
   inS("dunked(1)") :- dunk(0).
   inS("dunked(2)") :- dunk(1).
   notok :- ninS("armed(1)"),ninS("-armed(1)"), inS("armed(0)"), dunk(0).
   notok :- ninS("armed(2)"),ninS("-armed(2)"),inS("armed(1)"), dunk(1).
   inS("-armed(1)") :- inS("dunked(0)"), flush(0).
   inS("-armed(2)") :- inS("dunked(1)"), flush(1).
   notok :- ninS("armed(2)").

%% 2.2 -- fixed rules

% Skipped, see QBF Encoding

%%% 3. constraint
   :- not notok.
\end{alltt}
}


\section{
Integrated Program for Strategic Companies}
\label{app:stratcomp}

The integrated program for the strategic companies problem instance
in Section~\ref{sec:strategic}, $\mathit{SC_{strategic}}$ = $\mathit{SC_{guess}} \cup
\mathit{SC_{check}}'$,  is given below.
It has two answer sets
$S_1$ = \{\texttt{strat(barilla)}, \texttt{strat(saiwa)}, \texttt{strat(frutto)}, \ldots \}
and
$S_2$ = \{\texttt{strat(barilla)}, \texttt{strat(panino)}, \ldots \}
which correspond to the strategic sets as identified above.

\medskip

{\footnotesize
\begin{alltt}
%%%% GUESS PART
  company(barilla). company(saiwa). company(frutto). company(panino).
  prod_by(pasta,barilla,saiwa).  prod_by(tomatoes,frutto,barilla).
  prod_by(wine,barilla,barilla). prod_by(bread,saiwa,panino).
  contr_by(frutto,barilla,saiwa,barilla).

%% Guess Program: Not necessarily minimal

  strat(X) v -strat(X) :- company(X).
   :-  prod_by(X,Y,Z), not strat(Y), not strat(Z).
   :-  contr_by(W,X,Y,Z), not strat(W),
       strat(X), strat(Y), strat(Z).

%%%% REWRITTEN CHECK PART (after grounding)
%% 1. Create dynamically the facts for the program:

% smaller :- -strat1(X).
  lit(h,"smaller",1). atom("smaller","smaller").
  lit(p,"-strat1(saiwa)",1).
  lit(h,"smaller",2). atom("smaller","smaller").
  lit(p,"-strat1(panino)",2).
  lit(h,"smaller",3). atom("smaller","smaller").
  lit(p,"-strat1(frutto)",3).
  lit(h,"smaller",4). atom("smaller","smaller").
  lit(p,"-strat1(barilla)",4).

% strat1(X) v -strat1(X) :- strat(X).
  lit(h,"strat1(saiwa)",5) :- strat(saiwa).    atom("strat1(saiwa)","strat1(saiwa)").
  lit(h,"-strat1(saiwa)",5) :- strat(saiwa).   atom("-strat1(saiwa)","strat1(saiwa)").
  lit(h,"strat1(panino)",6) :- strat(panino).  atom("strat1(panino)","strat1(panino)").
  lit(h,"-strat1(panino)",6) :- strat(panino). atom("-strat1(panino)","strat1(panino)").
lit(h,"strat1(frutto)",7) :- strat(frutto).    atom("strat1(frutto)","strat1(frutto)").
lit(h,"-strat1(frutto)",7) :- strat(frutto).   atom("-strat1(frutto)","strat1(frutto)").
lit(h,"strat1(barilla)",8) :- strat(barilla).  atom("strat1(barilla)","strat1(barilla)").
lit(h,"-strat1(barilla)",8) :- strat(barilla). atom("-strat1(barilla)","strat1(barilla)").
\end{alltt}

\noindent\texttt{\% For constraints, critical negative literals need to be represented (cf.}
\textbf{OPT\begin{math}_{mod}\end{math}}\texttt{)}

\begin{alltt}
% :- prod_by(X,Y,Z), not strat1(Y), not strat1(Z).
  lit(n,"strat1(saiwa)",10) :- prod_by(bread,saiwa,panino).
  lit(n,"strat1(panino)",10) :- prod_by(bread,saiwa,panino).
  lit(n,"strat1(frutto)",11) :- prod_by(tomatoes,frutto, barilla).
  lit(n,"strat1(barilla)",11) :- prod_by(tomatoes,frutto, barilla).
  lit(n,"strat1(barilla)",12) :- prod_by(wine,barilla,barilla).
  lit(n,"strat1(barilla)",13) :- prod_by(pasta,barilla,saiwa).
  lit(n,"strat1(saiwa)",13) :- prod_by(pasta,barilla,saiwa).

% :- contr_by(W,X,Y,Z), not strat1(W), strat1(X), strat1(Y), strat1(Z).
  lit(n,"strat1(frutto)",14) :- contr_by(frutto,barilla,saiwa,saiwa).

%% 2. Optimized meta-interpreter

%% 2.1 -- program dependent part
  inS("smaller") :- inS("-strat1(saiwa)").
  inS("smaller") :- inS("-strat1(panino)").
  inS("smaller") :- inS("-strat1(frutto)").
  inS("smaller") :- inS("-strat1(barilla)").
  notok :- ninS("strat1(saiwa)"),ninS("-strat1(saiwa)"),strat(saiwa).
  notok :- ninS("strat1(panino)"),ninS("-strat1(panino)"),strat(panino).
  notok :- ninS("strat1(frutto)"),ninS("-strat1(frutto)"),strat(frutto).
  notok :- ninS("strat1(barilla)"),ninS("-strat1(barilla)"),strat(barilla).
  notok :- ninS("smaller").
  notok :- ninS("strat1(saiwa)"),ninS("strat1(panino)").
  notok :- ninS("strat1(frutto)"),ninS("strat1(barilla)").
  notok :- ninS("strat1(barilla)").
  notok :- ninS("strat1(barilla)"),ninS("strat1(saiwa)").
  notok :- inS("strat1(barilla)"),inS("strat1(saiwa)"),ninS("strat1(frutto)").

%% 2.2 -- fixed rules

% Skipped, see QBF Encoding

%%% 3. constraint
   :- not notok.
\end{alltt}
}

\section{Ad Hoc Encoding for Quantified Boolean Formulas}
\label{app:adhoc-qbf}

The ad hoc encoding in \cite{leon-etal-2002-dlv} for evaluating a QBF
of form $F = \exists x_1 \cdots \exists x_m \forall y_1 \cdots
\forall y_n\,\Phi$, where $\Phi = c_1 \vee \cdots \vee c_k $ is a propositional
formula over
$x_1,\ldots,x_m, y_1,\ldots,y_n$ in 3DNF, i.e. each
$c_i = a_{i,1} \wedge \cdots \wedge a_{i,3}$ and $|a_{i,j}| \in
\{x_1,\ldots,$ $x_m,$ $y_1\ldots,$ $y_n\}$,
represents $F$ by the following facts:
\begin{itemize}
\item \texttt{exists($x_i$).} for each existential variable $x_i$;
\item \texttt{forall($y_j$).} for each universal variable $y_j$; and
\item \texttt{term($p_1, p_2, p_3, q_1, q_2, q_3$).}
for each disjunct
$c_j = l_{i,1} \land l_{i,2} \land l_{i,3}$ in $\Phi$, where
(i) if $l_{i,j}$ is a positive atom $v_k$, then $p_j = v_k$,
otherwise $p_j$= $\mathtt{true}$, and
(ii) if $l_{i,j}$ is a negated atom $\neg v_k$, then $q_i=v_k$,
otherwise $q_i$ = $\mathtt{false}$. For example, $term(x_1,\mathtt{true}, y_4,
\mathtt{false}, y_2, \mathtt{false})$, encodes the term $x_1\land \neg y_2
\land y_4$.
\end{itemize}
For instance, our sample instance from Section~\ref{sec:QBF}
$$
\exists x_0x_1\forall y_0y_1
(\tneg{x_0} \wedge \tneg{y_0}) \vee
(y_0 \wedge \tneg{x_0}) \vee
(y_1 \wedge x_0 \wedge \tneg{y_0}) \vee
(y_0 \wedge \tneg{x_1} \wedge \tneg{y_0})
$$
would be encoded by the following facts:
{\tt\small
\begin{alltt}
   exists(x0).  exists(x1).  forall(y1).  forall(y2).
   term(true,true,true,x0,y0,false).
   term(y0,true,true,x0,false,false).
   term(y1,x0,true,y0,false,false).
   term(y0,true,true,x1,y0,false).
\end{alltt}}
\noindent
These facts are conjoined with the following facts and rules:
{\small
\begin{alltt}
   t(true). f(false).
   t(X) \vel f(X) :- exists(X).
   t(Y) \vel f(Y) :- forall(Y).
             w :- term(X,Y,Z,Na,Nb,Nc),t(X),t(Y),t(Z),
                  f(Na),f(Nb),f(Nc).
          t(Y) :- w, forall(Y).
          f(Y) :- w, forall(Y).
               :- not w.
\end{alltt}}
The guessing part ``initializes'' the logical constants $\mathtt{true}$ and
$\mathtt{false}$ and
chooses a witnessing assignment $\sigma$ to the variables in $X$,
which leads to an answer set $M_G$ for this part. The more tricky
checking part then tests whether $\phi[X/ \sigma(X)]$ is a tautology,
using a saturation technique similar to our meta-interpreter.

\section{Ad Hoc Encodings for Strategic Companies}
\label{app:adhoc-sc}

The first ad hoc encoding for Strategic Companies in \cite{eite-etal-2000c}, $ad hoc_1$, solves the problem
in a surprisingly elegant way by the following two rules conjoined to
the facts representing the $\mathit{prod\_by}$ and $\mathit{contr\_by}$
relations:
{\small
\begin{alltt}
   strat(Y) \vel strat(Z) :- prod\_by(X,Y,Z).
              strat(W) :- contr\_by(W,X,Y,Z), strat(X),\ strat(Y),\ strat(Z).
\end{alltt}
}
Here, the minimality of answer sets plays together with the first
rule generating candidate strategic sets and the second rule enforcing
the constraint on the  controls relation. It constitutes a
sophisticated example of intermingled guess and check.
Howewer, this succinct encoding relies very much on the fixed number of
producing and controlling companies; an extension to arbitrarily many producers and controllers
seems not to be as easy as in our separate guess and check programs from
Section~\ref{sec:strategic}.

The second ad hoc encoding from \cite{eite-etal-2000c}, $ad hoc_2$,
strictly separates the guess and checking parts, and uses the
following rules and constraints:
{\small
\begin{alltt}
   strat(X) \vel -strat(X) :- company(X).
     :- prod\_by(X,Y,Z), not strat(Y), not strat(Z).
     :- contr\_by(W,X,Y,Z), not strat(W), strat(X), strat(Y), strat(Z).
     :- not min(X), strat(X).
     :- strat'(X,Y), -strat(Y).
     :- strat'(X,X).
   min(X) \vel strat'(X,Y) \vel strat'(X,Z) :- prod\_by(G,Y,Z),strat(X).
   min(X) \vel strat'(X,C) :- contr\_by(C,W,Y,Z), strat(X),
                           strat'(X,W), strat'(X,Y), strat'(X,Z).
            strat'(X,Y) :- min(X), strat(X), strat(Y), X{\kuneq}Y.
\end{alltt}
}

Informally, the first rule and the first two constraints generate a candidate
strategic set, whose minimality is checked by the remainder of the
program. For a detailed explanation, we refer to
\cite{eite-etal-2000c}.

\bibliographystyle{plain}

\begin{thebibliography}{10}

\bibitem{bald-gelf-2003}
Marcello Balduccini and Michael Gelfond.
\newblock {Diagnostic reasoning with A-Prolog}.
\newblock {\em {Journal of the Theory and Practice of Logic Programming}},
  3:425--461, July/September 2003.

\bibitem{bara-2002}
Chitta Baral.
\newblock {\em {Knowledge Representation, Reasoning and Declarative Problem
  Solving}}.
\newblock Cambridge University Press, 2002.

\bibitem{bene-dech-94}
R.~Ben-Eliyahu and R.~Dechter.
\newblock {Propositional Semantics for Disjunctive Logic Programs}.
\newblock {\em Annals of Mathematics and Artificial Intelligence}, 12:53--87,
  1994.

\bibitem{cado-etal-97}
Marco Cadoli, Thomas Eiter, and Georg Gottlob.
\newblock {Default Logic as a Query Language}.
\newblock {\em {IEEE Transactions on Knowledge and Data Engineering}},
  9(3):448--463, May/June 1997.

\bibitem{cima-rove-2000}
A.~Cimatti and M.~Roveri.
\newblock {Conformant planning via symbolic model checking}.
\newblock {\em {Journal of Artificial Intelligence Research}\/}~{\em 13},
  305--338, 2000.


\bibitem{dant-etal-01}
Evgeny Dantsin, Thomas Eiter, Georg Gottlob, and Andrei Voronkov.
\newblock {Complexity and Expressive Power of Logic Programming}.
\newblock {\em ACM Computing Surveys}, 33(3):374--425, 2001.

\bibitem{delg-etal-01}
Jim Delgrande, Torsten Schaub, and Hans Tompits.
\newblock {plp: A Generic Compiler for Ordered Logic Programs}.
\newblock In Thomas Eiter, Wolfgang Faber, and Miros{\l}aw Truszczy{\'n}ski,
  editors, {\em {Proceedings of the 6th International Conference on Logic
  Programming and Nonmonotonic Reasoning (LPNMR-01)}}, number 2173 in LNCS,
  pages 411--415. Springer, 2001.

\bibitem{eite-etal-2000c}
Thomas Eiter, Wolfgang Faber, Nicola Leone, and Gerald Pfeifer.
\newblock {Declarative Problem-Solving Using the DLV System}.
\newblock In Jack Minker, editor, {\em {Logic-Based Artificial Intelligence}},
  pages 79--103. Kluwer Academic Publishers, 2000.

\bibitem{eite-etal-2002a}
Thomas Eiter, Wolfgang Faber, Nicola Leone, and Gerald Pfeifer.
\newblock {Computing Preferred Answer Sets by Meta-Interpretation in Answer Set
  Programming}.
\newblock {\em {Journal of the Theory and Practice of Logic Programming}},
  3:463--498, July/September 2003.

\bibitem{eite-etal-2001e}
Thomas Eiter, Wolfgang Faber, Nicola Leone, Gerald Pfeifer, and Axel Polleres.
\newblock {A Logic Programming Approach to Knowledge-State Planning, II: the
  {\small DLV}$^{\cal K}$ System}.
\newblock {\em {Artificial Intelligence}}, 144(1--2):157--211, March 2003.

\bibitem{eite-etal-2001d}
Thomas Eiter, Wolfgang Faber, Nicola Leone, Gerald Pfeifer, and Axel Polleres.
\newblock {A Logic Programming Approach to Knowledge-State Planning: Semantics
  and Complexity}.
\newblock {\em {ACM Transactions on Computational Logic}}, 5(2), April 2004.

\bibitem{eite-etal-2002-tplp}
Thomas Eiter, Michael Fink, Giuliana Sabbatini, and Hans Tompits.
\newblock {On Properties of Update Sequences Based on Causal Rejection}.
\newblock {\em {Journal of the Theory and Practice of Logic Programming}},
  2(6):721--777, 2002.

\bibitem{eite-etal-97f}
Thomas Eiter, Georg Gottlob, and Heikki Mannila.
\newblock {Disjunctive Datalog}.
\newblock {\em {ACM Transactions on Database Systems}}, 22(3):364--418,
  September 1997.

\bibitem{dlv-web}
Wolfgang Faber and Gerald Pfeifer.
\newblock {\tt DLV} homepage, {since 1996}.
\newblock \url{http://www.dlvsystem.com/}.

\bibitem{gelf-lifs-91}
M.~Gelfond and V.~Lifschitz.
\newblock {Classical Negation in Logic Programs and Disjunctive Databases}.
\newblock {\em {New Generation Computing}}, 9:365--385, 1991.

\bibitem{gelf-2002}
Michael Gelfond.
\newblock {Representing Knowledge in A-Prolog}.
\newblock In Antonis~C. Kakas and Fariba Sadri, editors, {\em {Computational
  Logic. Logic Programming and Beyond}}, number 2408 in LNCS, pages 413--451.
  Springer, 2002.

\bibitem{gold-bodd-96}
R.~Goldman and M.~Boddy.
\newblock {Expressive Planning and Explicit Knowledge}.
\newblock In {\em {Proceedings AIPS-96}}, pages 110--117. AAAI Press, 1996.

\bibitem{grec-etal-2001}
G.~Greco, S.~Greco, and E.~Zumpano.
\newblock A logic programming approach to the integration, repairing and
  querying of inconsistent databases.
\newblock In {\em Proceedings 17th International Conference on Logic
  Programming (ICLP 2001)}, number 2237 in {Lecture Notes in AI (LNAI)}, pages
  348--364. Springer Verlag, 2001.

\bibitem{janh-2000}
Tomi Janhunen.
\newblock Comparing the expressive powers of some syntactically restricted
  classes of logic porgrams.
\newblock In John Lloyd, Veronica Dahl, Ulrich Furbach, Manfred Kerber,
  Kung-Kiu Lau, Catuscia Palamidessi, Lu{\'\i}s~Moniz Pereira, Yehoshua Sagiv,
  and Peter~J. Stuckey, editors, {\em Computational Logic - CL 2000, First
  International Conference, Proceedings}, number 1861 in {Lecture Notes in AI
  (LNAI)}, pages 852--866, London, UK, July 2000. Springer Verlag.

\bibitem{janh-2001}
Tomi Janhunen.
\newblock On the effect of default negation on the expressiveness of
  disjunctive rules.
\newblock In Thomas Eiter, Wolfgang Faber, and Miros{\l}aw Truszczy\'nski,
  editors, {\em {Logic Programming and Nonmonotonic Reasoning --- 6th
  International Conference, LPNMR'01, Vienna, Austria, September 2001,
  Proceedings}}, number 2173 in {Lecture Notes in AI (LNAI)}, pages 93--106.
  Springer Verlag, September 2001.

\bibitem{janh-etal-2000}
Tomi Janhunen, Ilkka Niemel{\"a}, Patrik Simons, and Jia-Huai You.
\newblock {Partiality and Disjunctions in Stable Model Semantics}.
\newblock In Anthony~G. Cohn, Fausto Giunchiglia, and Bart Selman, editors,
  {\em {Proceedings of the Seventh International Conference on Principles of
  Knowledge Representation and Reasoning (KR~2000), April 12-15, Breckenridge,
  Colorado, USA}}, pages 411--419. Morgan Kaufmann Publishers, Inc., 2000.

\bibitem{lee-lifs-2003}
Joohyung Lee and Vladimir Lifschitz.
\newblock {Loop Formulas for Disjunctive Logic Programs}.
\newblock In {\em Proceedings of the Nineteenth International Conference on
  Logic Programming (ICLP-03)}. Springer Verlag, December 2003.

\bibitem{leon-etal-2002-dlv}
Nicola Leone, Gerald Pfeifer, Wolfgang Faber, Thomas Eiter, Georg Gottlob,
  Simona Perri, and Francesco Scarcello.
\newblock {The DLV System for Knowledge Representation and Reasoning}, 2004.
\newblock To appear. Available via \url{http://www.arxiv.org/ps/cs.AI/0211004}.

\bibitem{leon-etal-2001}
Nicola Leone, Riccardo Rosati, and Francesco Scarcello.
\newblock {Enhancing Answer Set Planning}.
\newblock In Alessandro Cimatti, H{\'e}ctor Geffner, Enrico Giunchiglia, and
  Jussi Rintanen, editors, {\em {IJCAI-01 Workshop on Planning under
  Uncertainty and Incomplete Information}}, pages 33--42, August 2001.

\bibitem{lifs-turn-94}
V.~Lifschitz and H.~Turner.
\newblock {Splitting a Logic Program}.
\newblock In Pascal {Van Hentenryck}, editor, {\em {Proceedings of the 11th
  International Conference on Logic Programming (ICLP'94)}}, pages 23--37,
  Santa Margherita Ligure, Italy, June 1994. MIT Press.

\bibitem{lifs-2002}
Vladimir Lifschitz.
\newblock {Answer Set Programming and Plan Generation}.
\newblock {\em {Artificial Intelligence}}, 138:39--54, 2002.

\bibitem{lin-zhao-2002}
Fangzhen Lin and Yuting Zhao.
\newblock {ASSAT: Computing Answer Sets of a Logic Program by SAT Solvers}.
\newblock In {\em {Proceedings of the Eighteenth National Conference on
  Artificial Intelligence (AAAI-2002)}}, Edmonton, Alberta, Canada, 2002. AAAI
  Press / MIT Press.

\bibitem{mare-remm-2003}
Victor~W. Marek and Jeffrey~B. Remmel.
\newblock {On the expressibility of stable logic programming}.
\newblock {\em {Journal of the Theory and Practice of Logic Programming}},
  3:551--567, November 2003.

\bibitem{mare-trus-99}
Victor~W. Marek and Miros{\l}aw Truszczy{\'n}ski.
\newblock {Stable Models and an Alternative Logic Programming Paradigm}.
\newblock In K.~Apt, V.~W. Marek, M.~Truszczy{\'n}ski, and D.~S. Warren,
  editors, {\em {The Logic Programming Paradigm -- A 25-Year Perspective}},
  pages 375--398. Springer Verlag, 1999.

\bibitem{mcde-87}
Drew McDermott.
\newblock {A Critique of Pure Reason}.
\newblock {\em {Computational Intelligence}}, 3:151--237, 1987.
\newblock Cited in \cite{cima-rove-2000}.

\bibitem{niem-99}
Ilkka Niemel{\"a}.
\newblock {Logic Programming with Stable Model Semantics as Constraint
  Programming Paradigm}.
\newblock {\em {Annals of Mathematics and Artificial Intelligence}},
  25(3--4):241--273, 1999.

\bibitem{papa-85}
Christos~H. Papadimitriou.
\newblock A note on the expressive power of prolog.
\newblock {\em Bulletin of the {EATCS}}, 26:21--23, 1985.

\bibitem{papa-94}
Christos~H. Papadimitriou.
\newblock {\em {Computational Complexity}}.
\newblock Addison-Wesley, 1994.

\bibitem{poll-2003}
Axel Polleres.
\newblock {\em {Advances in Answer Set Planning}}.
\newblock PhD thesis, {Institut f{\"u}r Informationssysteme, Technische
  Universit{\"a}t Wien}, {Wien, {\"O}sterreich}, 2003.

\bibitem{asp-2001}
Alessandro Provetti and Son~Tran Cao, editors.
\newblock {\em {Proceedings AAAI 2001 Spring Symposium on Answer Set
  Programming: Towards Efficient and Scalable Knowledge Representation and
  Reasoning}}, Stanford, CA, March 2001. AAAI Press.

\bibitem{przy-89b}
T.~Przymusinski.
\newblock {On the Declarative and Procedural Semantics of Logic Programs}.
\newblock {\em {Journal of Automated Reasoning}}, 5(2):167--205, 1989.

\bibitem{przy-91}
Teodor~C. Przymusinski.
\newblock {Stable Semantics for Disjunctive Programs}.
\newblock {\em {New Generation Computing}}, 9:401--424, 1991.

\bibitem{saka-inou-2003}
Chiaki Sakama and Katsumi Inoue.
\newblock {An abductive framework for computing knowledge base updates}.
\newblock {\em {Journal of the Theory and Practice of Logic Programming}},
  3:671--713, November 2003.

\bibitem{schl-95}
J.S. Schlipf.
\newblock {The Expressive Powers of Logic Programming Semantics}.
\newblock {\em {Journal of Computer and System Sciences}}, 51(1):64--86, 1995.
\newblock Abstract in Proc.\ PODS 90, pp.\ 196--204.

\bibitem{turn-2002}
Hudson Turner.
\newblock {Polynomial-Length Planning Spans the Polynomial Hierarchy}.
\newblock In Sergio Flesca, Sergio Greco, Giovambattista Ianni, and Nicola
  Leone, editors, {\em Proceedings of the 8th European Conference on Logics in
  Artificial Intelligence (JELIA)}, number 2424 in {Lecture Notes in Computer
  Science}, pages 111--124, Cosenza, Italy, September 2002. Springer.

\end{thebibliography}

\vspace*{-2ex}
\newcommand{\SortNoOp}[1]{}

\end{document}